\documentclass[lettersize,journal]{IEEEtran}
\usepackage{caption}
\usepackage{amsmath,amsfonts}
\usepackage{algorithmic}
\usepackage{algorithm}
\usepackage{array}
\usepackage{textcomp}
\usepackage{stfloats}
\usepackage{url}
\usepackage{verbatim}
\usepackage{graphicx}
\usepackage{cite}
\usepackage{hyperref}
\usepackage{amsmath,amssymb,amsfonts}
\usepackage{multirow}
\usepackage{makecell}
\usepackage[table]{xcolor}
\usepackage{threeparttable}

\begin{document}

\title{Multi-level Asymmetric Contrastive Learning for Medical Image Segmentation Pre-training}
\author{Shuang Zeng, Lei Zhu, Xinliang Zhang, Micky C Nnamdi, Wenqi Shi, J Ben Tamo, Qian Chen, Hangzhou He, Lujia Jin, Zifeng Tian, Qiushi Ren, Zhaoheng Xie, Yanye Lu*

\thanks{Shuang Zeng, Lei Zhu, Xinliang Zhang, Qian Chen, Hangzhou He, Lujia Jin, Zifeng Tian, Qiushi Ren, Zhaoheng Xie and Yanye Lu are with the Department of Biomedical Engineering, Peking University, Beijing, China and Institute of Medical Technology, Peking University Health Science Center, Peking University, Beijing, China, also with the National Biomedical Imaging Center, Peking University, Beijing, China, and also with the Institute of Biomedical Engineering, Peking University Shenzhen Graduate School, Shenzhen, China.}
\thanks{Micky C Nnamdi, J Ben Tamo are with school of Electrical and Computer Engineering, Georgia Institute of Technology, Atlanta, GA, USA.}
\thanks{Wenqi Shi are with Peter O’Donnell Jr. School of Public Health at UT Southwestern Medical Center (UTSW), Dallas, TX, USA.}
\thanks{*The corresponding author: Yanye Lu, (e-mail: yanye.lu@pku.edu.cn)}}

\markboth{IEEE Transactions On Image Processing}%
{Shell \MakeLowercase{\textit{et al.}}: A Sample Article Using IEEEtran.cls for IEEE Journals}


\maketitle

\begin{abstract}
Medical image segmentation is a fundamental yet challenging task due to the arduous process of acquiring large volumes of high-quality labeled data from experts. Contrastive learning offers a promising but still problematic solution to this dilemma. Firstly existing medical contrastive learning strategies focus on extracting image-level representation, which ignores abundant multi-level representations. Furthermore they underutilize the decoder either by random initialization or separate pre-training from the encoder, thereby neglecting the potential collaboration between the encoder and decoder. To address these issues, we propose a novel multi-level asymmetric contrastive learning framework named MACL for volumetric medical image segmentation pre-training. Specifically, we design an asymmetric contrastive learning structure to pre-train encoder and decoder simultaneously to provide better initialization for segmentation models. Moreover, we develop a multi-level contrastive learning strategy that integrates correspondences across feature-level, image-level, and pixel-level representations to ensure the encoder and decoder capture comprehensive details from representations of varying scales and granularities during the pre-training phase. Finally, experiments on 8 medical image datasets indicate our MACL framework outperforms existing 11 contrastive learning strategies. {\itshape i.e.} Our MACL achieves a superior performance with more precise predictions from visualization figures and 1.72\%, 7.87\%, 2.49\% and 1.48\% Dice higher than previous best results on ACDC, MMWHS, HVSMR and CHAOS with 10\% labeled data, respectively. And our MACL also has a strong generalization ability among 5 variant U-Net backbones. Our code will be released at \url{https://github.com/stevezs315/MACL}.
\end{abstract}


\section{Introduction}
\IEEEPARstart{M}{edical} image segmentation, defined as the partition of the entire image into different tissues, organs or other biologically relevant structures, plays a critical role in the computer-aided diagnosis paradigm. Although models with fully supervised training can dramatically improve the performance of segmentation tasks, it remains a challenge to assemble such large amount of annotated medical image datasets due to the extensive and burdensome
annotation effort and the requirement of expertise.
Meanwhile, substantial unlabeled image data from modalities 
such as CT and MRI are generated every day worldwide. Therefore, it is highly 
desirable to propose some methods which can alleviate the following 
requirement: leveraging numerous unlabeled data to pre-train models and achieve high performance with limited annotations.

Self-supervised learning (SSL) offers a promising solution to this dilemma: it provides a pre-training strategy that relies only on unlabeled data to obtain a suitable initialization for training downstream tasks with limited annotations. In recent years, SSL methods have demonstrated great success not only in the downstream analysis of natural images but also in medical images.
As a particular variant of SSL, contrastive learning (CL) \cite{chen2020simple, he2020momentum} has shown great success in learning image-level features from large-scale unlabeled data, significantly reducing annotation costs. 
In CL framework setting, an encoder is initially pre-trained on unlabeled data to extract valuable image representations. Subsequently, the pre-trained encoder serves as a well-suited initialization for training supervised downstream tasks, such as classification, object detection 
and segmentation, which can be fine-tuned into an accurate model with additional 
modules ({\itshape e.g.} a decoder for segmentation), 
even with limited labeled data. 
Additionally, CL framework requires an appropriate {\itshape contrastive loss} \cite{chen2020simple} to effectively attract similar pair representations ({\itshape a.k.a.} positive pairs) and repel dissimilar pair representations ({\itshape a.k.a.} negative pairs). This is predicated on the observation that different augmentations of the same image should yield similar representations, while representations derived from distinct images ought to differ.

\begin{figure*}[t]
  \centering
  \includegraphics[width=\textwidth]{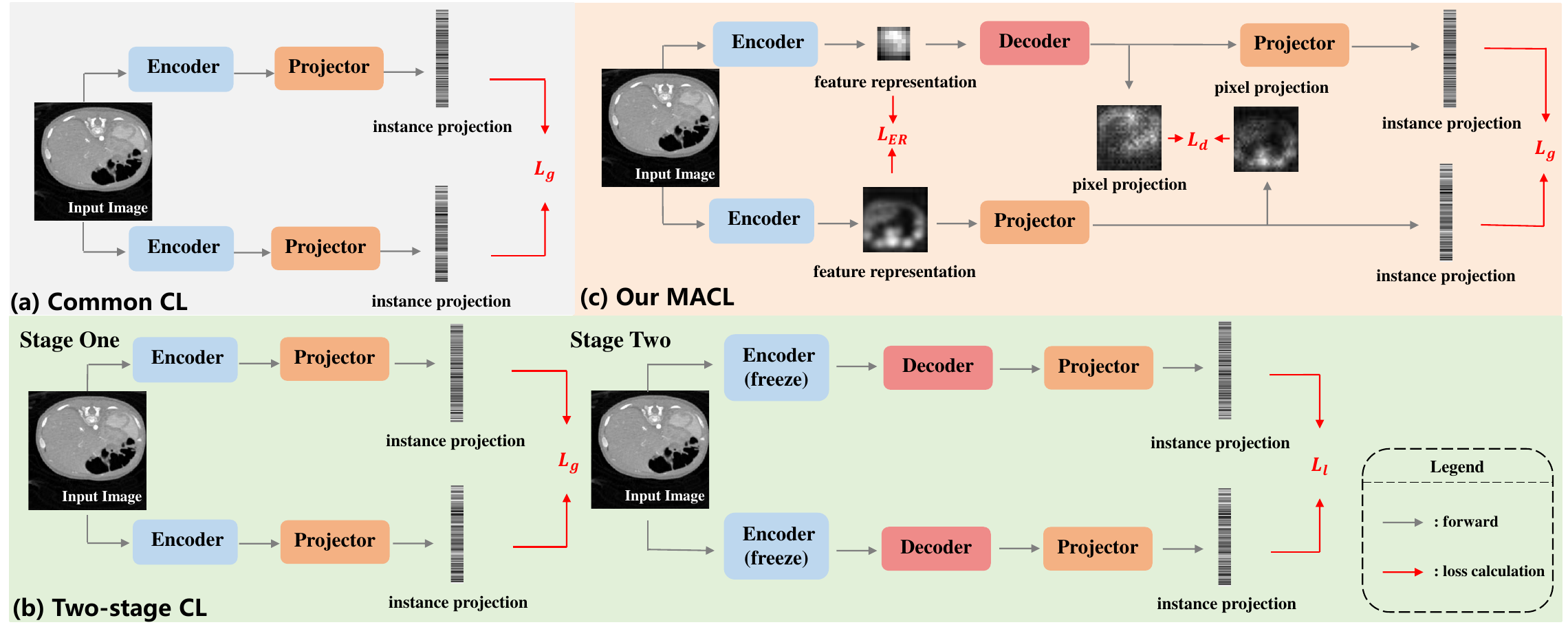}
  \caption{Comparison of different CL frameworks for medical images: (a) Common CL frameworks used for medical image segmentation are symmetric and similar to SimCLR with a global contrastive loss $\mathcal{L}_{g}$ for optimization. (b) Two-stage CL framework pre-trains the encoder with global contrastive loss $\mathcal{L}_{g}$ and decoder with local contrastive loss $\mathcal{L}_{l}$ in separate stage which ignores the collaboration between the encoder and decoder. (c) Our proposed MACL framework is asymmetric with an additional partial decoder for pre-training and integrates multi-level CL strategy including feature-level equivariant regularization, image-level and pixel-level CL to get better initialization of both encoder and decoder for downstream segmentation tasks.}
  \label{Overview}
\end{figure*}

In terms of medical image segmentation, several medical CL frameworks \cite{zeng2021positional,DeSD,DiRA,liu2021simtriplet, GCL} in Figure \ref{Overview}(a) and (b) have shown exciting results, which inspire us to excavate the potential of medical CL. Specifically, they pre-train a U-Net encoder along with an image-level projector and image-level contrastive loss to extract inter-image discrimination from unlabeled upstream datasets. And then they combine the well pre-trained encoder with a randomly initialized decoder for fine-tuning the segmentation task on downstream datasets with a small fraction of labeled data. However, despite its success, there are still some important issues to be solved when applying CL to this specific medical image segmentation task: (1) Medical image segmentation aims to predict the class label for each pixel within one image, which focuses on intra-image discrimination, instead of inter-image discrimination. In this sense, it poses demands for more distinctive pixel-level representation learning, which will be more appropriate for medical image segmentation. However, those medical CL works \cite{zeng2021positional,DeSD,DiRA,liu2021simtriplet, GCL} focus on extracting image-level projection output from the encoder along with an image-level projector, which do not exploit the details of both feature-level projection from the encoder and pixel-level projection from the decoder. Therefore, these pre-trained models can be sub-optimal for segmentation task due to the discrepancy between image-level projection and pixel-level projection and the missing of the implicit feature-level consistency. (2) The symmetric encoder-decoder architecture is widely used for fully supervised medical image segmentation. And due to the downsampling layers of encoder, deconvolutional layers of decoder and skip connections, the encoder-decoder structure can not only effectively extract multi-level representations but also preserve intricate details and edge information within medical images, hence facilitating more precise segmentation of subtle substructures such as tumors and cells. 
Such an efficient collaborative work paradigm inspires us that incorporating a decoder together with an encoder for CL pre-training should enhance the performance of downstream segmentation task. However, those proposed medical CL frameworks do not make full use of the decoder because their decoder is either randomly initialized and trained during downstream fine-tuning \cite{zeng2021positional, DeSD, DiRA, liu2021simtriplet} or pre-trained separately from the encoder, ignoring the collaboration between the encoder and decoder during pre-training \cite{GCL}. Specifically, the feature map from the encoder is highly compressed with fewer details which is suitable for global image-level CL while the feature map from the decoder is fine-grained with more texture which aligns well with local pixel-level CL. That means pre-training encoder and decoder simultaneously can allow the model to capture features at different levels and complement each other.

To fill the gap, we propose a novel \textbf{M}ulti-level \textbf{A}symmetric \textbf{CL} framework named MACL shown in Figure \ref{Overview}(c) by introducing an asymmetric CL structure and a multi-level CL strategy to realize one-stage encoder-decoder synchronous pre-training.  
Different from the existing works, there are mainly two characteristics of our proposed framework: 1) We propose a multi-level CL strategy ({\itshape i.e. image-level, feature-level, pixel-level}) from global to local to make sure luxuriant details from multi-scale and multi-granularity representations can be learned by the encoder and decoder during pre-training. 2) 
We incorporate a decoder into normal {\itshape one-encoder} CL framework to realize the synchronous encoder-decoder pre-training and further make an adaptive modification to the decoder which finally forms an asymmetric CL structure. Specifically, instead of directly introducing a symmetric encoder-decoder structure to both of the two branches, we introduce an encoder with a partial decoder in the dominant branch while only an encoder in the auxiliary branch. With such an asymmetric structure, our MACL can delightfully reduce the computation complexity of the model and guarantee sufficient negative pairs for CL. According to our experimental results, our proposed MACL outperforms other 11 methods across 8 medical image datasets. In summary, our main contributions are threefold:

(1) A multi-level contrastive learning strategy is designed to take the correspondence among image-level, pixel-level and feature-level projections respectively into account to make sure multi-level representations can be learned by the encoder and decoder during pre-training.

(2) In order to get better initialization for the entire downstream segmentation model, our proposed MACL adopts an asymmetric encoder-decoder CL structure, which introduces a partial decoder into the one-stage CL framework and pre-trains the encoder and decoder simultaneously.

(3) State-of-the-art results have been achieved across 8 datasets from both CT and MRI for multi-organ segmentation and ROI-based segmentation compared with other 11 CL methods. Further, we also conduct experiments to verify the generalization ability of our proposed MACL and ablation studies to validate the effectiveness of our proposed MACL.  

\section{Related Work}

\label{sec2}
\subsection{Medical Image Segmentation}

In recent years, deep learning techniques, especially convolutional neural networks (CNNs), have emerged as powerful tools for medical image segmentation.
Among those designs, FCN \cite{long2015fully}, UNet\cite{ronneberger2015u} and DeepLab \cite{chen2018encoder} function as three milestones. These models and their variants provide a stable and consistent baselines for segmentation tasks.
Among the most notable architectures, U-Net \cite{ronneberger2015u} stands out for its effective use of encoder-decoder structures and skip connections, enabling precise delineation of anatomical structures. Building upon U-Net's success, attention U-Net \cite{oktay2018attention} incorporates attention gates (AGs) to the skip connections to implicitly learn to suppress irrelevant regions in the input image while highlighting the regions of interest for the segmentation task.
Similarly, in order to bridge the semantic gaps for an accurate medical image segmentation, UCTransNet \cite{wang2022uctransnet} proposes a CTrans module to replace the original skip connection. This module consists of two sub-modules: CCT, which conducts multi-scale channel cross-fusion with a transformer, and CCA, which incorporates channel-wise cross-attention. Moreover, RollingUNet \cite{Liu_Zhu_Liu_Yu_Chen_Gao_2024} combines MLP with U-Net to efficiently fuse local features and long-range dependencies. Additionally, the development of Kolmogorov-Arnold Networks (KANs) \cite{liu2024kan} offered superior interpretability and efficiency through the utilization of a series of nonlinear, learnable activation functions. Therefore, UKAN \cite{li2024ukanmakesstrongbackbone} integrated KANs into UNet framework, augmenting its capacity for non-linear modeling while also improving model interpretability.
For 3D medical image segmentation, nnUNet \cite{isensee2021nnu} accomplishes searching suitable hyper-parameters including augmentations, pre-processing and post-processing techniques, etc. Its pipeline and code-base provide a strong baseline for a broad scope of tasks.


\subsection{Contrastive Learning}
As a particular variant of SSL, CL has shown great success in learning image-level features from large-scale unlabeled data which greatly reduces annotation costs, hence serving as a vital role in video frame prediction, semantic segmentation, object detection, and more.
The core idea of CL is to attract the positive sample 
pairs and repulse the negative sample pairs through optimizing a model 
with InfoNCE \cite{oord2018representation} loss. In practice, CL methods 
benefit from a large number of negative samples \cite{he2020momentum, chen2020simple}. MoCo \cite{he2020momentum} introduces a dynamic memory bank to maintain a queue of negative 
samples. 
SimCLR \cite{chen2020simple} employs a substantial batch size to facilitate the coexistence of a significant number of negative samples within the current batch.
Beyond the aforementioned 
classic training framework, SwAV \cite{caron2020unsupervised} incorporates online clustering into 
Siamese network and proposes a new `multi-crop' augmentation strategy that 
mixes the views of different resolutions. BYOL \cite{grill2020bootstrap} introduces a slow-moving 
average network and shows that CL can be effective without 
any negative samples. By introducing stop-gradient, SimSiam \cite{chen2021exploring} shows that 
simple Siamese networks can learn meaningful representations even without 
negative pairs, large batches or momentum encoders. 

\subsection{Contrastive Learning for Medical Image Segmentation}

In the medical imaging domain, substantial efforts of contrastive learning \cite{hervella2020multi,DeSD,DiRA,GCL,liu2021simtriplet,zeng2021positional} have been devoted to incorporating unlabeled data to improve network performance due to the limited data and annotations.
Researchers mainly focus on the design of siamese networks and construction of contrastive pairs to pre-train the model for several downstream tasks such as medical image classification, segmentation, registration and retrieval. Specifically, in terms of segmentation task, for the design of siamese networks, DeSD \cite{DeSD} reformulates CL in a deep self-distillation manner to improve the representation quality of both shallow and deep sub-encoders to address the issue that SSL may suffer from the weak supervision at shallow layers. DiRA \cite{DiRA} designs a novel self-supervised learning framework by uniting discriminative, restorative and adversarial learning in a unified manner to glean complementary visual information from unlabeled medical images. For the construction of contrastive pairs, SimTriplet \cite{liu2021simtriplet} proposes a {\itshape triplet-shape} CL framework which maximizes both intra-sample and inter-sample similarities via triplets from positive pairs without using negative samples to take advantage of multi-view nature of medical images. GCL \cite{GCL} proposes a 
{\itshape partition-based} contrasting strategy which leverages structural similarity across volumetric medical images to divide unlabeled samples into positive and negative pairs and formulates a local version of contrastive loss to learn distinctive local 
representation used for per-pixel segmentation. PCL \cite{zeng2021positional} further proposes a {\itshape position-based} CL framework to generate contrastive 
data pairs by leveraging the position information in volumetric medical images which 
can alleviate the problem that simple CL methods and GCL still 
introduce a lot of false negative pairs and result in degraded segmentation quality 
due to the circumstance where different medical images may have similar structures or 
organs.

\section{Methodology}\label{sec3}
This section first defines how to pre-train a model with CL framework for medical image segmentation. Then we propose a novel asymmetric CL structure which can pre-train both encoder and decoder in a single stage. Then, the multi-level contrastive learning strategy of our MACL is discussed to consider multi-level representations. Finally, the experimental setup including datasets and implementation details is described.

\subsection{Problem Definition}

Given an input image $\boldsymbol{X} \in \mathbb{R}^{C_x \times H_x \times W_x}$, medical image segmentation aims to classify pixels in the image with a segmentation network, where $H_x \times W_x$ is the resolution of the input image, $C_x$ denotes the channels of the input image. To achieve this purpose, the segmentation network needs an encoder $e(\cdot)$ to extract multi-level features and then a decoder $d(\cdot)$ is used to fuse the features into $\boldsymbol{Z}$ to recover image details:
\begin{equation}
    \boldsymbol{Z} = d(e(\boldsymbol{X})) = d(\{\boldsymbol{X}^1, \cdots, \boldsymbol{X}^L\})
\end{equation}
where $\boldsymbol{X}^i$ represents the $i_{th}$-level feature. $L$ is the number of encoder layers. $e(\cdot)$ and $d(\cdot)$ are two hypothetic function that can be approximated by the network with learning parameter $\theta_{e}$ and $\theta_{d}$. This fused feature $\boldsymbol{Z}$ is finally classified by a segmentation head to output segmentation map.

Except for optimizing $e(\cdot)$ and $d(\cdot)$ from scratch, CL focuses on pre-training them with numerous unlabeled data to provide suitable initialization before supervised learning for downstream segmentation task. However, as shown in Figure \ref{Overview}(a), most existing CL frameworks \cite{liu2021simtriplet,zeng2021positional,GCL,DeSD,DiRA} used for medical image segmentation are symmetric 
and similar to SimCLR \cite{chen2020simple}, which only pre-trains an encoder $e(\cdot)$ with a global contrastive loss $\mathcal{L}_{g}$ among image-level sample pairs:
\begin{equation}
    \underset{\mathbf{\theta}_{e}}{\arg \min} \quad \mathcal{L}_{g}  (e(\boldsymbol{\tilde{X}}), e(\boldsymbol{\hat{X}}))
\end{equation}
where $\boldsymbol{\tilde{X}}$ and $\boldsymbol{\hat{X}}$ are two random augmentations of the input image  $\boldsymbol{X}$. The decoder structure $d(\cdot)$ is not considered in the above pre-training process, which ignores the importance of decoders to the downstream segmentation task. To solve this problem, as shown in Figure \ref{Overview}(b), GCL\cite{GCL} proposes a two-stage CL framework which utilizes two pre-training stages to optimize $e(\cdot)$ with a global contrastive loss $\mathcal{L}_{g}$ and $d(\cdot)$ with a local contrastive loss $\mathcal{L}_{l}$ respectively:

\begin{equation}
\left\{\begin{array}{l}
\underset{\mathbf{\theta}_e}{\arg \min } \quad \mathcal{L}_{g} (e(\boldsymbol{\tilde{X}}), e(\boldsymbol{\hat{X}})), \quad \text{\itshape{in stage one}}\\
\underset{\mathbf{\theta}_d}{\arg \min } \quad \mathcal{L}_{l} (d(e(\boldsymbol{\tilde{X}})), d(e(\boldsymbol{\hat{X}}))), \quad \text{\itshape{in stage two}}\\
\end{array}\right.
\end{equation}
where $\boldsymbol{\tilde{X}}$ and $\boldsymbol{\hat{X}}$ are two random augmentations of the input image  $\boldsymbol{X}$, the parameters of $e(\cdot) $ are freezed in stage two. Although this strategy takes the decoder $d(\cdot)$ into consideration, the two-stage training process ignores the collaboration between the encoder and decoder during pre-training, which makes the decoder not able to learn image-level representation well due to the absence of global contrastive loss.

Based on the above issues, in order to realize the synchronous training of $e(\cdot)$ and $d(\cdot)$ to learn multi-level representations, we propose a novel one-stage multi-level asymmetric CL framework shown in Figure \ref{Overview}(c). Our MACL framework pre-trains the encoder $e(\cdot)$ and decoder $d(\cdot)$ simultaneously with both image-level global contrastive loss $\mathcal{L}_{g}$ and pixel-level dense contrastive loss $\mathcal{L}_{d}$ to learn multi-level feature representations:

\begin{equation}
    \underset{\mathbf{\theta}_{e}, \mathbf{\theta}_{d}}{\arg \min} \quad \mathcal{L}_{g}  (d(e(\boldsymbol{\tilde{X}})), d(e(\boldsymbol{\hat{X}})))
    + \mathcal{L}_{d}  (d(e(\boldsymbol{\tilde{X}})), d(e(\boldsymbol{\hat{X}})))
\end{equation}
where $\boldsymbol{\tilde{X}}$ and $\boldsymbol{\hat{X}}$ are two random augmentations of the input image  $\boldsymbol{X}$. With our proposed MACL, both the encoder and decoder can synchronously learn multi-level representations and get better initialization.

\begin{figure*}[t]
  \centering
  \includegraphics[width=\textwidth]{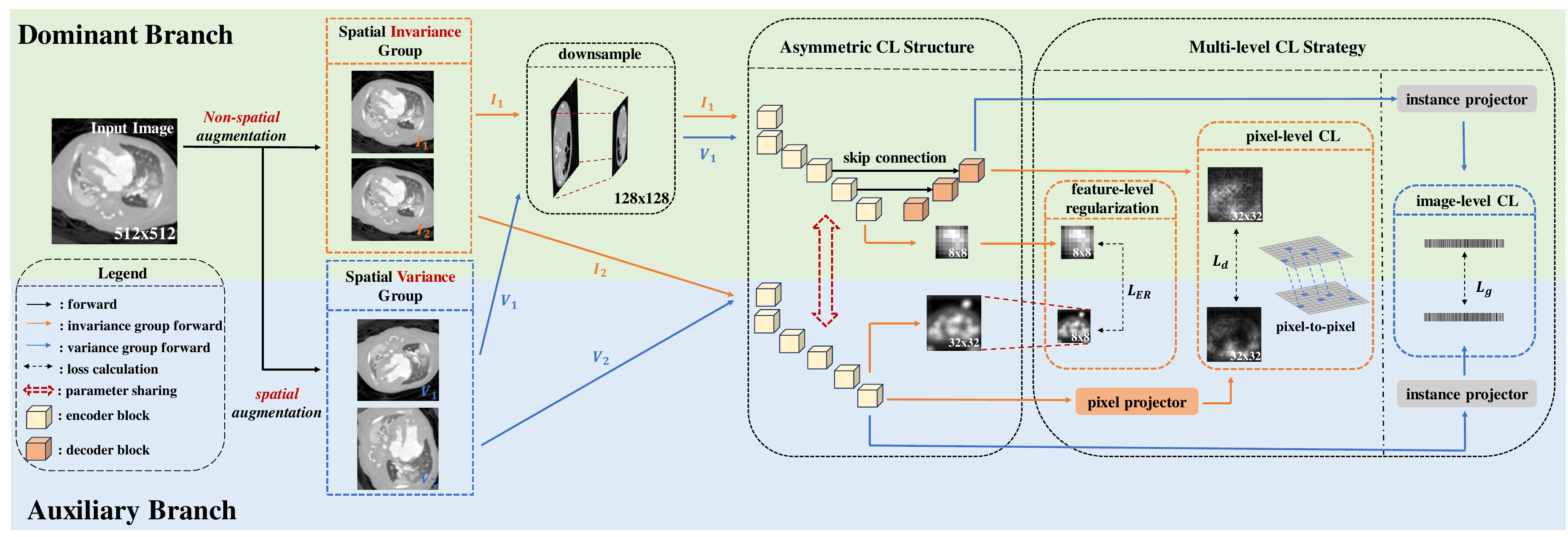}
  \caption{Overview of our proposed MACL framework. The input image is propagated to two branches: the dominant branch and auxiliary branch. In the dominant branch, after augmentation and downsampling, the input image is fitted into the encoder to get feature-level representation, and then propagated to the decoder and projector to get image-level and pixel-level projections; while in the auxiliary branch, it is the same process but no decoder will be used. The feature-level representations from two branches will be used in equivariant regularization loss, while image-level and pixel-level projections will be used in global and dense contrastive loss, respectively to achieve multi-level contrastive learning.}
  \label{JCL}
\end{figure*}

\subsection{Asymmetric Contrastive Learning Structure}

In order to realize the one-stage synchronous training of encoder $e(\cdot)$ and decoder $d(\cdot)$ for better assisting the downstream segmentation tasks, our MACL elaborates an asymmetric CL structure as shown in Figure \ref{JCL}.
This proposed structure contains two branches: (1) {\itshape dominant branch} which pre-trains both encoder and decoder simultaneously to get better initialization for downstream segmentation tasks and (2) {\itshape auxiliary branch} which serves as a contrastive counterpart to assist the dominant branch for pre-training. 

Before being fed into the asymmetric CL structure, two different settings for data augmentations will be applied to the input image $\boldsymbol{X}$: (1) One set is only non-spatial augmentations $T_{fix}$ which will be used for CL involving pixel-wise comparison including feature-level and pixel-level CL; (2) The other set is the combination of non-spatial and spatial augmentations $T_{var}$ which is suitable for image-level CL not involving pixel-wise comparison:
\begin{equation}
\begin{gathered}
  \{\boldsymbol{I_1}, \boldsymbol{I_2}\} = T_{fix}(\boldsymbol{X}), \quad
  \{\boldsymbol{V_1}, \boldsymbol{V_2}\} = T_{var}(\boldsymbol{X})
\end{gathered}
\end{equation}
where $\boldsymbol{I_1}, \boldsymbol{I_2}$ are augmented images of spatial invariance group and $\boldsymbol{V_1}, \boldsymbol{V_2}$ are augmented images of spatial variance group.
\subsubsection{Auxiliary Branch}

The auxiliary branch keeps in line with that of common CL strategies: an encoder $e(\cdot)$ followed by a shallow multi-layer perceptron (MLP) projection head $\tilde{g}(\cdot)$ \cite{chen2020simple} (but including an
image-level projector $\tilde{g}_{ins}(\cdot)$ and pixel-level projector $\tilde{g}_{pix}(\cdot)$) and no decoder. 

Specifically, the augmented image with pixel position fixed $\boldsymbol{I_2}$ is propagated to the encoder $e(\cdot)$ to get feature-level representation $\tilde{\boldsymbol{Y}}$ and then propagated into the pixel-level projector $\tilde{g}_{pix}(\cdot)$ to get pixel-level projection $\tilde{\boldsymbol{Z}}^{l} \in \mathbb{R}^{B\times C \times H \times W}$ (where B denotes batch size, C denotes channels of the projection, $H\times W$ denotes the spatial size of the projection):
\begin{equation}
\begin{gathered}
  \tilde{\boldsymbol{Y}} = e((\boldsymbol{I_2})), \quad \tilde{\boldsymbol{Z}}^{l} = \tilde{g}_{pix}(\tilde{\boldsymbol{Y}}) 
\end{gathered}
\end{equation}
While the augmented image with pixel position changed $\boldsymbol{V_2}$ is propagated into the encoder $e(\cdot)$ and the image-level projector $\tilde{g}_{ins}(\cdot)$ to get image-level projection $\tilde{\boldsymbol{Z}}^{g} \in \mathbb{R}^{B\times C}$:
\begin{equation}
    \begin{gathered}
  \tilde{\boldsymbol{Z}}^{g} = \tilde{g}_{ins}(e(\boldsymbol{V}_2))\\
  \end{gathered}
\end{equation}

\subsubsection{Dominant Branch}

Different from the auxiliary branch, the dominant branch introduces a decoder $d(\cdot)$ after the encoder $e(\cdot)$ for synchronous pre-training to dig more information from feature-level representation which is beneficial to the downstream segmentation tasks. The decoder $d(\cdot)$ is also followed by a MLP projection head $g(\cdot)$ (including an image-level projector $g_{ins}(\cdot)$ and pixel-level projector $g_{pix}(\cdot)$ as well). Moreover, due to the introduction of the decoder in this branch, the size of feature maps from two branches will be different if inputs are same. Hence, we add downsampling $down(\cdot)$ into data augmentation to align features which can also guarantee sufficient negative sample pairs due to its small computation complexity. 

Specifically, the augmented image with pixel position fixed $\boldsymbol{I_1}$ is first downsampled, and then propagated to the encoder $e(\cdot)$ which shares the same parameters with that in the auxiliary branch to get the feature-level representation $\boldsymbol{Y}$ and finally propagated to the decoder $d(\cdot)$ to recover image details which also serves as the pixel-level projection $\boldsymbol{Z}^{l}$:
\begin{equation}
\begin{gathered}
  \boldsymbol{Y} = e(down(\boldsymbol{I}_1)), \quad \boldsymbol{Z}^{l} = d(\boldsymbol{Y}) 
\end{gathered}
\end{equation}
While the augmented image with pixel position changed $\boldsymbol{V_1}$ is downsampled, propagated to the encoder $e(\cdot)$ and the image-level projector $g_{ins}(\cdot)$ to get image-level projection $\boldsymbol{Z}^{g}$:
\begin{equation}
  \begin{gathered}
  \boldsymbol{Z}^{g} = g_{ins}(e(down(\boldsymbol{V}_1)))
  \end{gathered}
\end{equation}

With such a novel asymmetric network structure to guarantee the feature alignment between the representation output after $e(\cdot)$, $d(\cdot)$ in the dominant branch and that output after $e(\cdot)$ in the auxiliary branch, we can realize one-stage synchronous training of $e(\cdot)$ and $d(\cdot)$ with more negative sample pairs and less computation complexity for better assisting the downstream segmentation tasks.

\subsection{Multi-level Contrastive Learning Strategy}

Under the guarantee of our proposed asymmetric CL structure in Section B which pre-trains the encoder and decoder simultaneously,
we can optimize our MACL framework with multi-level contrastive learning strategy considering image-level, pixel-level and feature-level representations as follows:
\begin{equation}
\mathcal{L}_{MLC}=\lambda_1 \mathcal{L}_{g}+\lambda_2 \mathcal{L}_{d}+\lambda_{3} \mathcal{L}_{ER}
\end{equation}
where $\lambda_1, \lambda_2, \lambda_3$ are the weighting factors, $\mathcal{L}_{MLC}$ is the total multi-level contrastive loss, and $\mathcal{L}_{g}, \mathcal{L}_{d}, \mathcal{L}_{ER}$ will be discussed later. The overall pre-training is conducted in one-stage manner. We give the details of network training settings in Experimental Setup and carefully investigate the effectiveness of our MACL in different perspectives in Results and Discussions.


\subsubsection{Global Contrastive Learning}

The image-level representation is regulated by the global contrastive loss.
Specifically, for a set of $N$ randomly sampled slices: $\left\{X_i\right\}_{i=1 \ldots N}$, the corresponding mini-batch
consists of $2N$ samples after data augmentations, $\left\{\tilde{X}_i\right\}_{i=1 \ldots 2 N}$, in which $\tilde{X}_{2 i}$
and $\tilde{X}_{2i-1}$ are two random augmentations of $X_{i}$. $Z^{g}_{i}$, $\tilde{Z}_{i}^{g}$ represent the learned image-level projection of $\tilde{X}_{i}$ from
the dominant branch and auxiliary branch, respectively. Then the global contrastive loss can be formulated as:

\begin{equation}\label{global_CL}
  \mathcal{L}_{g}=\sum_{i=1}^{2 N} -\frac{1}{\left|\Omega_i^{+}\right|} \sum_{j \in \Omega_i^{+}} \log \frac{e^{s i m\left(Z_{i}^{g}, \tilde{Z}_{j}^{g}\right) / \tau}}{\sum_{k=1}^{2 N} \mathbb{I}_{i \neq k} \cdot e^{s i m\left(Z_{i}^{g}, \tilde{Z}_{k}^{g}\right) / \tau}}
  \end{equation}
where $\left|\Omega_i^{+}\right|$ is the set of indices of positive samples to $\tilde{X}_{i}$. $sim(\cdot, \cdot)$ is the cosine similarity function that
computes the similarity between two vectors in the representation space. $\tau$ is a temperature scaling parameter. $\mathbb{I}$ is an indicator function. Compared with the standard contrastive loss \cite{chen2020simple} that only
has one positive pair on the numerator for any sample $X_i$, we use the contrastive pair generation strategy in PCL \cite{zeng2021positional} to form {\itshape position-based} positive and negative pairs. 


\subsubsection{Dense Contrastive Learning}

The pixel-level representation is forced by our proposed dense contrastive loss which takes the correspondence between the pixel-level features into account and extends the original global contrastive loss to a dense paradigm. In detail, for a set of $N$ randomly sampled slices: $\left\{X_i\right\}_{i=1 \ldots N}$, the corresponding mini-batch
consists of $2N$ samples after data augmentation, $\left\{\tilde{X}_i\right\}_{i=1 \ldots 2 N}$, in which $\tilde{X}_{2 i}$
and $\tilde{X}_{2i-1}$ are two random augmentations of $X_{i}$. Different from the image-level projection 
$Z^{g}_{i}$ and $\tilde{Z}_{i}^{g}$, $Z^{l}_{i}$ and $\tilde{Z}_{i}^{l}$ are pixel-level projections generated by the pixel-level 
projector in the dominant and auxiliary branch, respectively, which are dense feature maps with a size of $S\times S$. $S$ denotes the spatial size of the generated dense feature maps.
For every two input images, we can form $S^2$ pixel-level projection pairs (positive or negative pairs depending on $\Delta$ {\itshape position}) according to the {\itshape position-based} contrastive pair generation strategy. So the dense contrastive loss can be defined as:

\begin{equation}\label{local_CL}
\scalebox{0.95}{
  $\mathcal{L}_{d}=\sum_{i=1}^{2 N} -\frac{1}{\left|\Omega_i^{+}\right|} \sum_{j \in \Omega_i^{+}} \frac{1}{S^2} \sum_s \log \frac{e^{s i m\left(Z_{i,s}^{l}, \tilde{Z}_{j,s}^{l}\right) / \tau}}{\sum_{k=1}^{2 N} \mathbb{I}_{i \neq k} \cdot e^{s i m\left(Z_{i,s}^{l}, \tilde{Z}_{k,s}^{l}\right) / \tau}}$}
\end{equation}
where $Z_{i,s}^{l}$ and $\tilde{Z}_{i,s}^{l}$ denote the $s_{th}$ output of $S^2$ pixel-level projection pairs of $\tilde{X}_i$ from the dominant branch and auxiliary branch, respectively. 

\subsubsection{Equivariant Regularization}

The feature-level representation is governed by equivariant regularization \cite{wang2020self}.
Specifically, due to the introduction of downsampling in the dominant branch, the feature representations from the two encoders are different in size, therefore, we incorporate equivariant regularization into our multi-level CL to apply consistency regularization on multi-scale feature representations to provide further self-supervision for network learning. The equivariant regularization can be formulated as:
\begin{equation}
  \mathcal{L}_{E R}=\|\boldsymbol{Y} - down(\tilde{\boldsymbol{Y}})\|_1
  \end{equation}
where $\boldsymbol{Y}$ and $\tilde{\boldsymbol{Y}}$ are the representations learned from the encoder in the dominant and auxiliary branch, respectively. We use L1 norm here to integrate consistency regularization on feature
representations from the encoders to provide further self-supervision for network learning.

\begin{table*}[t]
\centering
\begin{threeparttable}
\caption{Quantitative results of our MACL against other SOTA methods on 4 multi-organ segmentation datasets.}
\label{multiorgan}
\scriptsize
\setlength{\tabcolsep}{1pt}
\begin{tabular}{c|c|cccc|cccc|cccc|cccc}
\Xhline{1.0pt} 
 &  & \multicolumn{4}{c|}{ ACDC (100 patients) } & \multicolumn{4}{c|}{ MMWHS (20 patients) } & \multicolumn{4}{c|}{ HVSMR (10 patients) } & \multicolumn{4}{c}{ CHAOS (20 patients)} \\
 {Labeled}& {Methods} & DSC (\%) $\uparrow$ & JC (\%) $\uparrow$ & HD95 $\downarrow$ & ASD $\downarrow$ & DSC (\%) $\uparrow$ & JC (\%) $\uparrow$ & HD95 $\downarrow$ & ASD $\downarrow$ 
 & DSC (\%) $\uparrow$ & JC (\%) $\uparrow$ & HD95 $\downarrow$ & ASD $\downarrow$
 & DSC (\%) $\uparrow$ & JC (\%) $\uparrow$ & HD95 $\downarrow$ & ASD $\downarrow$
 \\
\Xhline{1.0pt} \multirow{12}*{10 \%}
& Scratch &$79.99$ &$67.39$ &$8.78$ &$3.82$ &$63.04$ &$48.89$ &$9.31$ &$3.39$ 
&$72.81$ &$59.02$ &$34.34$ &$10.18$ &$51.76$ &$37.51$ &$23.08$ &$9.90$ 
\\
& SimCLR \cite{chen2020simple} &$78.60$ &$65.47$ &$8.11$ &$3.36$ &$61.70$ &$46.59$ &$9.67$ &$3.53$ 
&$75.78$ &$62.34$ &$25.16$ &${\mathbf{6.22}}$ &$53.22$ &$38.11$ &$22.28$ &$9.04$
\\
& MoCo \cite{he2020momentum} &$81.34$ &$69.13$ &$8.12$ &$3.20$ &$57.54$ &$42.62$ &$11.33$ &$4.15$
&$76.20$ &$62.79$ &$27.45$ &$8.33$ &$53.14$ &$38.49$ &$21.45$ &$9.53$
\\
& BYOL \cite{grill2020bootstrap} &$77.55$ &$64.17$ &$9.05$ &$3.96$ &$63.75$ &$49.54$ &$9.34$ &$3.61$ 
&$73.57$ &$59.87$ &$29.79$ &$8.51$ &$50.45$ &$35.66$ &$24.20$ &$10.32$
\\
& SwAV \cite{caron2020unsupervised} &$82.11$ &$70.20$ &$\underline{6.35}$ &$\underline{2.46}$ &$64.27$ &$49.96$ &$10.31$ &$3.98$ 
&$\underline{76.28}$ &$\underline{62.91}$ &$24.09$ &$6.42$ &$49.87$ &$35.04$ &$19.75$ &$8.86$
\\
& SimSiam \cite{chen2021exploring} &$71.96$ &$57.01$ &$11.65$ &$4.94$ &$61.37$ &$46.51$ &$10.61$ &$3.89$
&$75.10$ &$61.47$ &$29.73$ &$8.72$ &$41.21$ &$28.87$ &$21.55$ &$10.20$
\\
& SimTriplet \cite{liu2021simtriplet} &$71.07$ &$56.07$ &$11.31$ &$5.20$ &$62.52$ &$47.81$ &$10.98$ &$4.10$ 
&$72.66$ &$58.48$ &$30.72$ &$8.65$ &$32.88$ &$23.14$ &$27.81$ &$12.90$
\\
& GCL \cite{GCL} &$84.86$ &$74.07$ &$7.07$ &$2.85$ &$\underline{71.41}$ &$\underline{58.58}$ &$\underline{7.27}$ &$\underline{2.55}$
&$75.70$ &$62.22$ &$28.15$ &$7.68$ &$58.78$ &$44.00$ &${\mathbf{15.28}}$ &$\underline{6.36}$
\\
& PCL \cite{zeng2021positional} &$\underline{84.91}$ &$\underline{74.16}$ &$6.54$ &$2.65$ &$65.98$ &$52.32$ &$8.07$ &$3.16$
&$76.12$ &$62.82$ &$24.50$ &$6.43$ &$\underline{63.61}$ &$\underline{48.23}$ &${16.15}$ &${6.79}$
\\
& DeSD \cite{DeSD} &$76.02$ &$62.27$ &$10.25$ &$4.15$ &$57.03$ &$42.30$ &$10.89$ &$4.36$
&$74.57$ &$61.16$ &$\underline{24.00}$ &$6.40$ &$61.93$ &${46.68}$ &$18.91$ &$8.39$
\\
& DiRA \cite{DiRA} &$83.14$ &$71.61$ &$6.73$ &$2.60$ &$57.73$ &$42.95$ &$10.92$ &$3.76$
&$74.87$ &$61.35$ &$27.91$ &$8.01$ &$44.30$ &$31.24$ &$24.63$ &$11.93$
\\
& \cellcolor{cyan!20!green!20!}MACL (Ours) &\cellcolor{cyan!20!green!20!}$\mathbf{86.63}$ &\cellcolor{cyan!20!green!20!}$\mathbf{76.72}$ 
&\cellcolor{cyan!20!green!20!}$\mathbf{5.23}$ 
&\cellcolor{cyan!20!green!20!}$\mathbf{1.73}$ 
&\cellcolor{cyan!20!green!20!}$\mathbf{79.28}$ 
&\cellcolor{cyan!20!green!20!}$\mathbf{67.19}$ 
&\cellcolor{cyan!20!green!20!}$\mathbf{5.91}$ 
&\cellcolor{cyan!20!green!20!}$\mathbf{1.98}$ 
&\cellcolor{cyan!20!green!20!}$\mathbf{78.77}$ 
&\cellcolor{cyan!20!green!20!}$\mathbf{66.05}$ 
&\cellcolor{cyan!20!green!20!}$\mathbf{22.38}$ 
&\cellcolor{cyan!20!green!20!}$\underline{6.25}$ 
&\cellcolor{cyan!20!green!20!}$\mathbf{65.09}$ 
&\cellcolor{cyan!20!green!20!}$\mathbf{51.12}$ 
&\cellcolor{cyan!20!green!20!}$\underline{18.95}$ 
&\cellcolor{cyan!20!green!20!}$7.40$
\\
\Xhline{1.0pt} 
\multirow{12}*{25 \%}
& Scratch &$88.99$ &$80.34$ &$\underline{3.93}$ &$\underline{1.24}$ &$85.14$ &$74.87$ &$4.66$ &$1.45$
&$81.26$ &$69.25$ &$20.63$ &$5.85$ &$66.71$ &$52.74$ &$12.32$ &$5.39$
\\
& SimCLR \cite{chen2020simple} &$88.55$ &$79.71$ &$4.53$ &$1.47$ &$86.14$ &$76.03$ &$4.59$ &$1.36$
&$80.53$ &$68.50$ &$23.42$ &$6.75$ &$66.15$ &$52.21$ &$11.76$ &$4.75$
\\
& MoCo \cite{he2020momentum} &$88.38$ &$79.39$ &$5.01$ &$1.55$ &$82.89$ &$71.46$ &$5.40$ &$1.69$
&$81.26$ &$69.18$ &$20.41$ &$5.64$ &$65.33$ &$50.93$ &$15.12$ &$6.40$
\\
& BYOL \cite{grill2020bootstrap} &$\underline{89.27}$ &$\underline{80.78}$ &$4.62$ &$1.64$ &$87.63$ &$78.43$ &$3.83$ &$1.20$
&$81.33$ &$69.30$ &$19.95$ &$5.80$ &$62.49$ &$48.31$ &$14.13$ &$5.71$
\\
& SwAV \cite{caron2020unsupervised} &$88.99$ &$80.34$ &$4.90$ &$1.78$ &$87.71$ &$78.61$ &$\underline{3.78}$ &$1.22$
&$81.30$ &$69.18$ &$\underline{17.24}$ &$\underline{4.62}$ &$68.12$ &$54.01$ &$12.71$ &$5.15$
\\
& SimSiam \cite{chen2021exploring} &$88.24$ &$79.18$ &$4.85$ &$1.71$ &$82.71$ &$71.19$ &$5.25$ &$1.78$
&$79.70$ &$67.30$ &$22.83$ &$6.31$ &$51.82$ &$39.11$ &$14.82$ &$6.86$
\\
& SimTriplet \cite{liu2021simtriplet} &$87.99$ &$78.80$ &$5.78$ &$2.00$ &$82.43$ &$70.82$ &$5.24$ &$1.73$
&$80.51$ &$68.24$ &$24.34$ &$7.17$ &$63.55$ &$49.30$ &$13.23$ &$5.98$
\\
& GCL \cite{GCL} &$88.55$ &$79.67$ &$4.22$ &$1.34$ &$\underline{88.01}$ &$\underline{79.04}$ &$4.02$ &$\underline{1.17}$
&$82.41$ &$70.80$ &$22.08$ &$6.59$ &$76.88$ &$63.48$ &$9.11$ &$3.70$
\\
& PCL \cite{zeng2021positional} &$88.97$ &$80.33$ &$4.17$ &$1.43$ &$86.65$ &$76.90$ &$4.23$ &$1.28$
&$\underline{83.13}$ &$\underline{71.70}$ &$20.28$ &$5.17$ &$\underline{78.98}$ &$\underline{66.08}$ &$\underline{8.87}$ &$\underline{3.32}$
\\
& DeSD \cite{DeSD} &$87.14$ &$77.47$ &$6.11$ &$2.19$ &$84.46$ &$73.66$ &$5.15$ &$1.70$
&$81.44$ &$69.44$ &$21.19$ &$5.57$ &$67.96$ &$54.28$ &$11.70$ &$4.90$
\\
& DiRA \cite{DiRA} &$88.92$ &$80.22$ &$4.24$ &$1.48$ &$86.68$ &$77.08$ &$4.46$ &$1.43$ 
&$81.56$ &$69.67$ &$20.85$ &$5.69$ &$64.00$ &$49.87$ &$13.47$ &$5.69$
\\
& \cellcolor{cyan!20!green!20!} MACL (Ours) 
&\cellcolor{cyan!20!green!20!}$\mathbf{89.64}$
&\cellcolor{cyan!20!green!20!}$\mathbf{81.40}$ 
&\cellcolor{cyan!20!green!20!}$\mathbf{3.59}$ 
&\cellcolor{cyan!20!green!20!}$\mathbf{1.20}$ 
&\cellcolor{cyan!20!green!20!}$\mathbf{89.90}$ 
&\cellcolor{cyan!20!green!20!}$\mathbf{81.92}$ 
&\cellcolor{cyan!20!green!20!}$\mathbf{3.68}$ 
&\cellcolor{cyan!20!green!20!}$\mathbf{1.01}$ 
&\cellcolor{cyan!20!green!20!}$\mathbf{83.68}$ 
&\cellcolor{cyan!20!green!20!}$\mathbf{72.47}$ 
&\cellcolor{cyan!20!green!20!}$\mathbf{17.03}$ 
&\cellcolor{cyan!20!green!20!}$\mathbf{4.14}$ 
&\cellcolor{cyan!20!green!20!}$\mathbf{79.88}$ 
&\cellcolor{cyan!20!green!20!}$\mathbf{67.45}$ 
&\cellcolor{cyan!20!green!20!}$\mathbf{8.07}$ 
&\cellcolor{cyan!20!green!20!}$\mathbf{3.05}$
\\
\Xhline{1.0pt} 
\multirow{1}*{100 \%}
& Scratch &$92.07$ &$85.38$ &$3.06$ &$1.07$ &$92.27$ &$85.79$ &$2.71$ &$0.76$
&$84.95$ &$74.33$ &$14.83$ &$3.56$ &$88.65$ &$79.84$ &$4.38$ &$1.62$
\\
\Xhline{1.0pt}
\end{tabular}
\begin{tablenotes}
    \item[1]{The up arrow next to the metric means a larger value is better, while the down arrow means a smaller value is better. The value in \textbf{bold} represents the best and the \underline{underlined} values represent the second best. Scratch means no pre-training.}
  \end{tablenotes}
\end{threeparttable}
\end{table*}

\begin{table*}[th]
\centering
\caption{Quantitative results of our proposed MACL against other SOTA methods on 4 ROI-based segmentation datasets.}
\scriptsize
\setlength{\tabcolsep}{1pt}
\begin{tabular}{c|c|cccc|cccc|cccc|cccc}
\Xhline{1.0pt} 
 &  & \multicolumn{4}{c|}{ Spleen (41 patients) } & \multicolumn{4}{c|}{ ISIC (2594 images) } & \multicolumn{4}{c|}{ Heart (20 patients) } & \multicolumn{4}{c}{ Hippocampus (260 patients) }\\
 {Labeled}& {Methods} & DSC (\%) $\uparrow$& JC (\%) $\uparrow$ & HD95 $\downarrow$ & ASD $\downarrow$ & DSC (\%) $\uparrow$ & JC (\%) $\uparrow$ & HD95 $\downarrow$ & ASD $\downarrow$ 
 & DSC (\%) $\uparrow$ & JC (\%) $\uparrow$ & HD95 $\downarrow$ & ASD $\downarrow$
 & DSC (\%) $\uparrow$ & JC (\%) $\uparrow$ & HD95 $\downarrow$ & ASD $\downarrow$
\\
\Xhline{1.0pt} \multirow{12}*{10 \%}
& Scratch &$66.63$ &$49.95$ &$12.58$ &$5.78$&$\underline{87.21}$&$\underline{77.32}$&$23.47$&$9.72$&$73.30$ &$57.85$ &$5.21$ &$1.82$ &$\underline{78.27}$ &$\underline{64.30}$ &$1.36$ &$0.51$ \\
& SimCLR \cite{chen2020simple} &$60.38$ &$43.24$ &$12.42$ &$4.67$&$85.85$&$75.21$&$23.81$&$10.32$ &$73.42$&$58.00$&$\mathbf{4.32}$&$\underline{1.40}$ &$75.91$ &$61.18$ &$1.48$ &$0.54$ \\
& MoCo \cite{he2020momentum} &$56.49$ &$39.36$ &$21.32$ &$9.35$&$85.61$&$74.84$&$24.56$&$9.95$ &$68.32$&$51.88$&$6.71$&$2.48$ &$78.27$ &$64.30$ &${\mathbf{1.27}}$ &$0.46$\\
& BYOL \cite{grill2020bootstrap}&$61.38$ &$44.28$ &$14.56$ &$6.32$&$87.00$&$77.00$&$24.20$&$10.28$&$70.34$ &$58.00$ &${{6.17}}$ &$2.16$ &$77.24$ &$62.92$ &$1.39$ &$0.49$  \\
& SwAV \cite{caron2020unsupervised} &$61.87$ &$44.79$ &$13.61$ &$6.53$&$85.85$&$75.20$&$24.19$&$10.45$&$72.31$ &$56.63$ &$5.95$ &$2.27$ &$76.00$ &$61.29$ &$1.47$ &$0.52$ \\
& SimSiam \cite{chen2021exploring} &$67.88$ &$51.37$ &$12.89$ &$3.65$&$83.02$&$70.97$&$30.04$&$13.70$&$58.56$ &$41.41$ &$10.71$ &$5.15$ &$73.45$ &$58.05$ &$1.39$ &$0.49$\\
& SimTriplet \cite{liu2021simtriplet} &$63.56$ &$46.59$ &$16.95$ &$7.54$&$83.01$&$70.96$&$29.27$&$13.03$&$57.76$ &$40.61$ &$8.85$ &$3.94$ &$74.80$ &$59.74$ &$1.31$ &$\underline{0.46}$ \\
& GCL \cite{GCL}&$\underline{73.40}$ &$\underline{57.97}$ &$11.79$ &$4.99$&$85.53$&$74.72$&$24.37$&$10.84$&$76.88$&$62.44$&${5.21}$&$1.73$&$78.14$ &$64.13$ &$1.32$ &$0.46$\\
& PCL \cite{zeng2021positional} &${71.22}$ &$55.30$ &$11.10$ &$4.97$&$86.64$&$76.43$&$24.41$&$10.50$ &$\underline{78.47}$ &$\underline{64.57}$ &$5.95$ &$2.14$ &$78.13$ &$64.11$ &$\underline{1.29}$ &$0.47$ \\
& DeSD \cite{DeSD} &$64.24$ &$47.31$ &$\underline{7.18}$ &${\mathbf{2.83}}$&$83.56$&$71.76$&$25.66$&$11.50$ &$77.55$ &$63.33$ &$5.80$ &$2.28$ &$74.48$ &$59.34$ &$1.64$ &$0.62$\\
& DiRA \cite{DiRA} &$66.23$ &$49.51$ &$20.96$ &$8.93$&$87.02$&$77.02$&$\underline{22.91}$&$\underline{9.28}$&$70.19$&$54.07$ &$\underline{4.76}$&${\mathbf{1.36}}$&$77.15$ &$62.80$ &$1.36$ &$0.50$\\
& \cellcolor{cyan!20!green!20!}MACL (Ours) 
&\cellcolor{cyan!20!green!20!}${\mathbf{80.93}}$ 
&\cellcolor{cyan!20!green!20!}${\mathbf{67.97}}$ 
&\cellcolor{cyan!20!green!20!}${\mathbf{6.93}}$ 
&\cellcolor{cyan!20!green!20!}$\underline{2.92}$
&\cellcolor{cyan!20!green!20!}${\mathbf{87.45}}$ 
&\cellcolor{cyan!20!green!20!}${\mathbf{77.70}}$ 
&\cellcolor{cyan!20!green!20!}${\mathbf{21.53}}$
&\cellcolor{cyan!20!green!20!}${\mathbf{8.91}}$ 
&\cellcolor{cyan!20!green!20!}${\mathbf{79.37}}$ 
&\cellcolor{cyan!20!green!20!}${\mathbf{65.79}}$ 
&\cellcolor{cyan!20!green!20!}$5.40$ 
&\cellcolor{cyan!20!green!20!}$1.99$ 
&\cellcolor{cyan!20!green!20!}${\mathbf{78.33}}$ 
&\cellcolor{cyan!20!green!20!}${\mathbf{64.38}}$ 
&\cellcolor{cyan!20!green!20!}${1.31}$ 
&\cellcolor{cyan!20!green!20!}${\mathbf{0.46}}$ \\
\Xhline{1.0pt} \multirow{12}*{25 \%}
& Scratch&$74.99$ &$59.98$ &$9.14$ &$4.33$&$88.44$&$79.28$&${\mathbf{16.32}}$&${\mathbf{6.41}}$&$78.35$&$64.41$ &$4.57$ &$1.33$ &$83.48$ &$71.65$ &$1.08$ &$0.35$\\
& SimCLR \cite{chen2020simple} &$81.75$ &$69.13$ &$7.05$ &$3.06$&$87.82$&$78.28$&$17.79$&$7.18$&$80.07$ &$66.76$ &${\mathbf{3.48}}$ &$0.98$ &$83.35$ &$71.46$ &$1.05$ &$0.36$\\
& MoCo \cite{he2020momentum}&$75.98$ &$61.26$ &$8.55$ &$4.09$&$87.53$&$77.83$&$17.87$&$7.01$&$75.35$&$60.45$&$5.60$&$2.11$ &$83.94$ &$72.33$ &$\underline{0.98}$ &$\underline{0.33}$\\
& BYOL \cite{grill2020bootstrap}&$75.15$ &$60.19$ &$6.34$ &$2.39$&$88.50$&$79.38$&$\underline{16.78}$&$\underline{6.54}$&$80.70$ &$67.65$ &$4.36$ &$1.16$ &$83.38$ &$71.50$ &$1.04$ &$0.35$\\
& SwAV \cite{caron2020unsupervised}&$80.40$ &$67.23$ &$\underline{5.52}$ &$\underline{2.34}$&$87.90$&$78.41$&$17.65$&$6.77$&$81.94$ &$69.40$ &$3.98$ &$1.22$ &$83.05$ &$71.02$ &$1.06$ &$0.35$\\
& SimSiam \cite{chen2021exploring}&$78.70$ &$64.88$ &$8.87$ &$4.56$&$87.29$&$77.45$&$19.31$&$7.52$&$71.65$ &$55.83$ &$\underline{3.74}$ &$1.17$ &$80.60$ &$67.51$ &$1.11$ &$0.37$\\
& SimTriplet \cite{liu2021simtriplet}&$78.81$ &$65.04$ &$8.84$ &$3.96$&$86.52$&$76.24$&$19.50$&$7.68$&$76.42$ &$61.83$ &$5.11$ &$1.72$ &$82.01$ &$69.51$ &$1.06$ &$0.35$\\
& GCL \cite{GCL} &$\underline{84.37}$ &$\underline{72.97}$ &$6.25$ &$2.52$&$\underline{88.52}$&$\underline{79.41}$&$17.55$&$7.06$&$82.30$ &$69.92$ &$3.76$ &$\underline{0.96}$ &$83.63$ &$71.87$ &$1.07$ &$0.37$\\
& PCL \cite{zeng2021positional} &$83.55$ &$71.74$ &$7.03$ &$2.96$&$88.26$&$78.99$&$18.04$&$7.27$&$\underline{82.69}$&$\underline{70.49}$&$3.78$&${\mathbf{0.93}}$ &$\underline{83.98}$ &$\underline{72.40}$ &${\mathbf{0.96}}$ &$0.33$\\
& DeSD \cite{DeSD} &$81.02$ &$68.10$ &$6.59$ &$3.28$&$87.34$&$77.52$&$20.29$&$8.41$&$81.50$ &$68.77$ &$4.45$ &$1.43$ &$81.81$ &$69.22$ &$1.08$ &$0.37$\\
& DiRA \cite{DiRA}&$80.80$ &$67.78$ &$7.95$ &$4.17$&$88.38$&$79.18$&$16.96$&$6.63$&$79.91$ &$66.54$ &$4.87$ &$1.35$ &$83.60$ &$71.83$ &$1.00$ &$0.33$ \\
&\cellcolor{cyan!20!green!20!}  MACL (Ours) &\cellcolor{cyan!20!green!20!}${\mathbf{85.82}}$ 
&\cellcolor{cyan!20!green!20!}${\mathbf{75.16}}$ 
&\cellcolor{cyan!20!green!20!}${\mathbf{4.85}}$ 
&\cellcolor{cyan!20!green!20!}${\mathbf{1.91}}$
&\cellcolor{cyan!20!green!20!}${\mathbf{88.96}}$
&\cellcolor{cyan!20!green!20!}${\mathbf{80.12}}$
&\cellcolor{cyan!20!green!20!}$17.61$
&\cellcolor{cyan!20!green!20!}$6.79$
&\cellcolor{cyan!20!green!20!}${\mathbf{82.93}}$ 
&\cellcolor{cyan!20!green!20!}${\mathbf{70.84}}$ 
&\cellcolor{cyan!20!green!20!}$3.85$ 
&\cellcolor{cyan!20!green!20!}$1.24$ 
&\cellcolor{cyan!20!green!20!}${\mathbf{84.13}}$ 
&\cellcolor{cyan!20!green!20!}${\mathbf{72.61}}$ 
&\cellcolor{cyan!20!green!20!}${0.99}$ 
&\cellcolor{cyan!20!green!20!}${\mathbf{0.32}}$ \\
\Xhline{1.0pt} \multirow{1}*{100 \%}
& Scratch &$89.92$ &$81.69$ &$4.59$ &$2.12$&$90.11$&$82.01$&$15.71$&$6.17$&$91.52$ &$84.36$ &$2.45$ &$0.61$ &$87.00$ &$77.00$ &$0.81$ &$0.27$ \\
\Xhline{1.0pt}
\end{tabular}
\label{ROI}
\end{table*}

\begin{figure*}[th]
  \centering
  \includegraphics[width=0.8\textwidth]{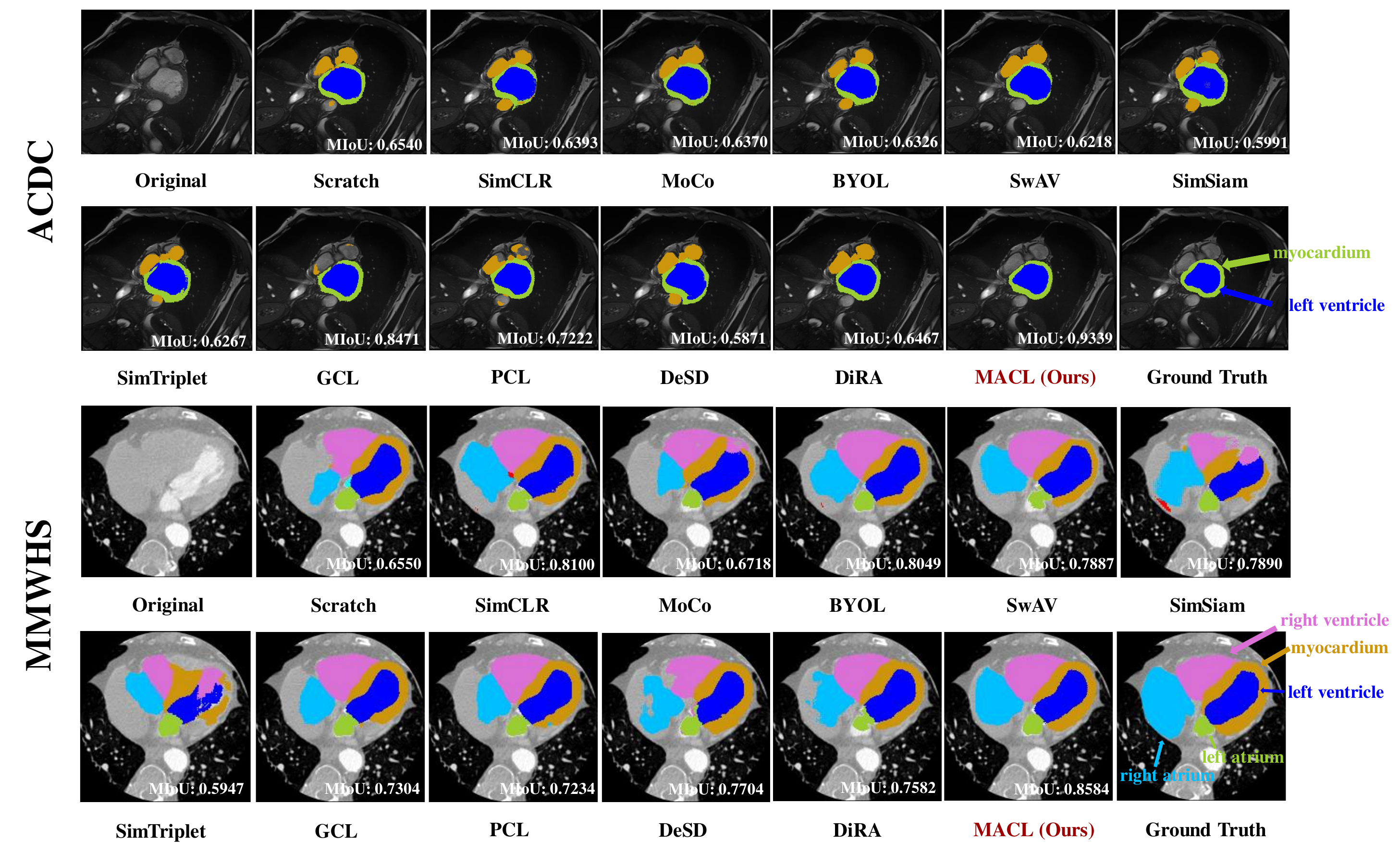}
  \caption{Visualization of multi-organ segmentation results on ACDC and MMWHS datasets. Our proposed MACL achieves a superior performance with a higher MIoU and more precise predictions of substructures across other 11 methods.}
  \label{Vis_1}
\end{figure*}

\begin{figure*}[th]
  \centering
  \includegraphics[width=0.8\textwidth]{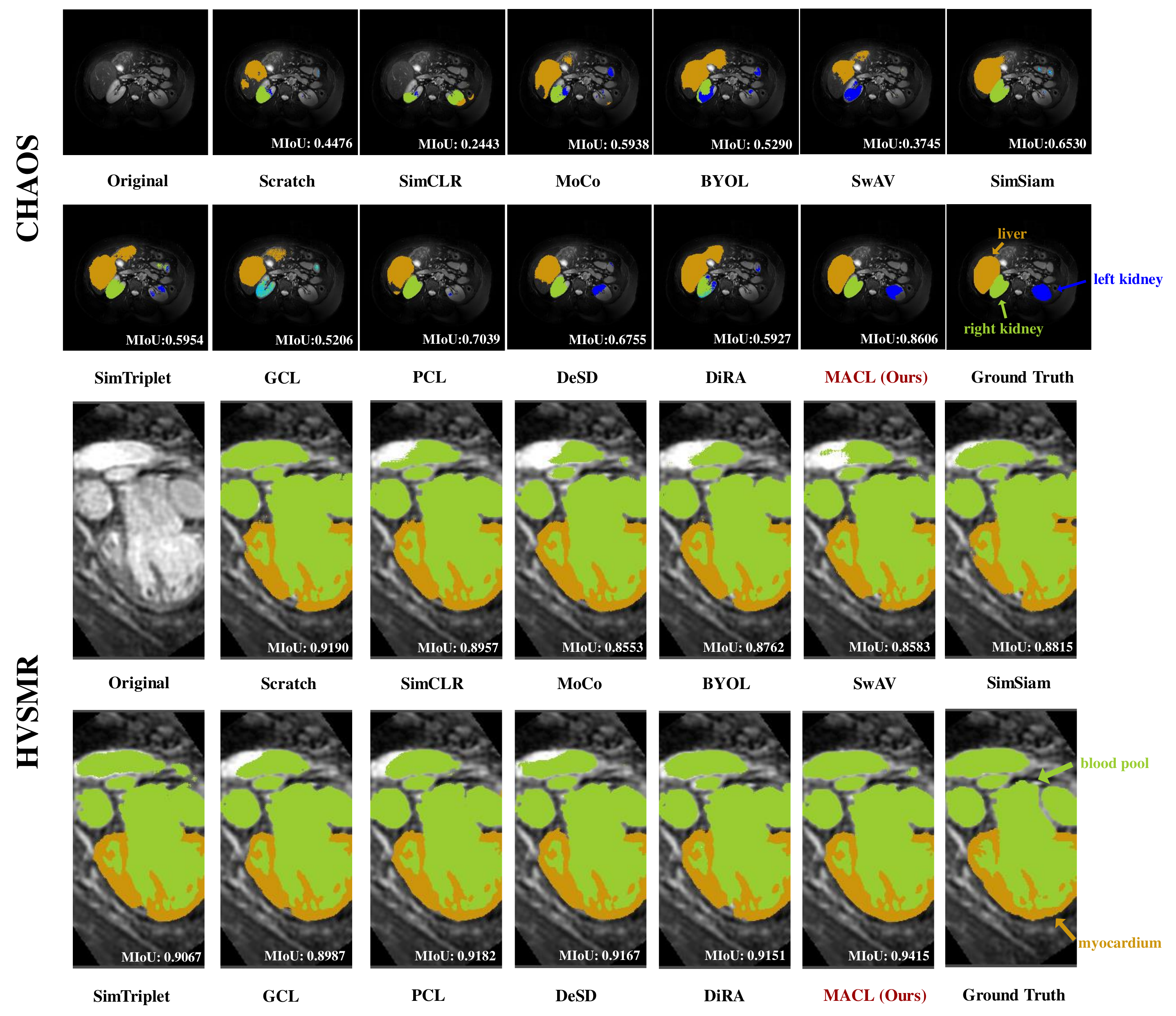}
  \caption{Visualization of multi-organ segmentation results on CHAOS and HVSMR datasets. Our proposed MACL achieves a superior performance with a higher MIoU and more precise predictions of substructures across other 11 methods.}
  \label{Vis_2}
\end{figure*}

\section{Experiments}\label{sec:experiments}

\subsection{Datasets}

We evaluate the performance of our proposed MACL on two different types of segmentation tasks: Multi-organ segmentation and ROI (regions-of-interest, such as tumor, skin lesion, etc.) based segmentation. We use 3 publicly available medical image datasets: CHD (17525 slices), BraTS2018 (39064 slices) and KiTS2019 (32332 slices) for pre-training and 8 datasets for fine-tuning and evaluation of performance.

\noindent{{\itshape \textbf{Multi-organ Segmentation:}}} 
\textbf{(1) The CHD dataset} \cite{PCL23} is a CT dataset that consists of 68 3D cardiac images captured by a Simens biograph 64 machine. The dataset covers 14 types of congenital heart disease and the segmentation labels include seven substructures: left ventricle (LV), right ventricle (RV), left atrium (LA), right atrium (RA), myocardium (Myo), aorta (Ao) and pulmonary artery (PA).
\textbf{(2) The ACDC dataset} \cite{ACDC} has 100 patients with 3D cardiac MRI images and the segmentation labels include three substructures: LV, RV and Myo. 
\textbf{(3) The MMWHS dataset} \cite{PCL24} consists of 20 cardiac CT and 20 MRI images and the annotations include the same seven substructures as the CHD dataset. And we use the 20 cardiac CT images for fine-tuning. 
\textbf{(4) The HVSMR dataset} \cite{PCL16} has 10 3D cardiac MRI images captured in an axial view on a 1.5T scanner. Manual annotations of blood pool and Myo are provided.
\textbf{(5) The CHAOS dataset} \cite{CHAOS2021} has totally 120 DICOM datasets from T1-DUAL in/out phase and T2-SPIR. And we choose T2-SPIR train dataset with 20 DICOM MRI images and corresponding labels of four regions: liver, left kidney, right kidney and spleen for fine-tuning.

\noindent{{\itshape \textbf{ROI-based Segmentation:}}} 
\textbf{(6) The BraTS2018 dataset} \cite{bakas2018identifying} contains a total of 351 patients scanned with T1, T1CE, T2, and Flair MRI volumes. And we choose 285 patients which are annotated with the segmentation labels including 3 tumor regions and 1 background region for pre-training. Notably, since there are no suitable MRI datasets aimed for multi-organ segmentation, we use the BraTS2018 dataset as the pre-training dataset for both multi-organ and ROI-based MRI segmentation downstream tasks.
\textbf{(7) The KiTS2019 dataset} \cite{heller2021state} consists of 210 patients, aimed for kidney and kidney tumor segmentation and we choose it as the pre-training dataset for ROI-based CT segmentation.   
\textbf{(8) The MSD dataset} \cite{antonelli2022medical} has 10 avaliable datasets, where each dataset has between one and three region-of-interest (ROI) targets. And we use three datasets: 
\textbf{Heart} (Task02, MRI dataset, 20 patients labeled with left atrium), \textbf{ Hippocampus} (Task04, MRI dataset, 260 patients labeled with anterior and posterior of hippocampus) and \textbf{Spleen} (Task09, CT datset, 41 patients labeled with spleen).
\textbf{(9) The ISIC2018 dataset} \cite{DBLP:journals/corr/abs-1902-03368} is a comprehensive collection of dermatoscopic images published by the International Skin Imaging Collaboration (ISIC). And for the lesion segmentation task, the dataset includes 2594 images with corresponding annotations of lesion skin region which were reviewed by dermatologists.

\subsection{Implementation Details}

We employ our MACL framework to pre-train a U-Net encoder and two-block decoder on CHD, BraTS and KiTS dataset respectively without using any human label. Then the pre-trained model is used as the initialization to fine-tune a U-Net segmentation network (that is to say we add the remaining decoder blocks so that the output of the network has the same dimensions as the input) with a small fraction (10\% / 25\%) of labeled samples of 8 downstream datasets, respectively. Data pre-processing follows the same setting as PCL \cite{zeng2021positional}. All the datasets are split into train set and test set in an 8:2 ratio.
Experiments of pre-training and fine-tuning are conducted with PyTorch on NVIDIA A100 80G GPUs. We find using the original images as spatial invariance group for calculating pixel-level CL loss has gained the best performance compared with using some non-spatial augmentations such as brightness, contrast, etc. Therefore, non-spatial augmentation is identity mapping, and spatial augmentations are translation, rotation and scale. In the pre-training stage, the weighted terms $\{\lambda_1, \lambda_2, \lambda_3\}$ are set to be $\{1.0, 0.5, 1.0\}$, temperature $\tau$ is set to be 0.1, the model is trained with 100 epochs, 16 batches per GPU, SGD optimizer and initial learning rate of 0.1, which is then decayed to 0 with the cosine scheduler on each training iterator. In the fine-tuning stage, we train the U-Net with cross-entropy loss for 100 epochs with Adam optimizer. The batch size per GPU is set to be 5, and initial learning rate is $5\times e^{-4}$ which is decayed with cosine scheduler to minimum learning rate $5 \times e^{-6}$. We evaluate the segmentation performance of our method along with other SOTA methods using 4 metrics, including DSC (Dice Similarity Coefficient), JC (Jaccard Coefficient), HD95 (95th percentile of the Hausdorff Distance) and ASD (Average Surface Distance) with respect to the predicted segmentation map and the label.

\subsection{Comparisons with SOTAs}
We compare the performance of our MACL with the scratch approach which does not use any pre-training strategy as well as other state-of-the-art baselines, including 1) SimCLR \cite{chen2020simple}, MoCo \cite{he2020momentum}, BYOL \cite{grill2020bootstrap}, SwAV \cite{caron2020unsupervised} and SimSiam \cite{chen2021exploring} originally used in natural imaging domain but modified with an encoder of UNet as the backbone of CL; 2) SimTriplet \cite{liu2021simtriplet}, GCL \cite{GCL}, PCL \cite{zeng2021positional}, DeSD \cite{DeSD}, DiRA \cite{DiRA} specially used in medical imaging domain. Notably, SimCLR, MoCo, PCL and DeSD are {\itshape symmetric one-stage} CL frameworks pre-training {\itshape only encoder}; GCL is a {\itshape symmetric two-stage} CL framework pre-training {\itshape encoder and decoder separately}; BYOL, SwAV, SimSiam and SimTriplet are {\itshape asymmetric one-stage} CL frameworks pre-training {\itshape only encoder}; DiRA is a {\itshape two-stage hybrid} SSL framework which has {\itshape symmetric} CL pre-training {\itshape only encoder}. However, our proposed MACL is an \textbf{asymmetric one-stage} CL framework pre-training \textbf{both encoder and decoder} together.  
All the experiments across different methods have the same dataset settings and partition, and all the methods have the same backbone structures. To make it clear, we summarize our comparative studies as ``upstream dataset (pre-training) $\rightarrow$ downstream dataset (fine-tuning)": for multi-organ segmentation task, it is CHD (CT) $\rightarrow$ ACDC/MMWHS, BraTS (MRI) $\rightarrow$ HVSMR/CHAOS; while for ROI-based segmentation task, it is KiTS (CT) $\rightarrow$ Spleen/ISIC, BraTS $\rightarrow$ Heart/Hippocampus. 
The quantitative results carried on multi-organ segmentation task are shown in Table \ref{multiorgan} while ROI-based segmentation task in Table \ref{ROI}.

\noindent{\textbf{Quantitative results:}} From Table \ref{multiorgan} and \ref{ROI}, we have the following observations: In general, benefited from our proposed asymmetric CL framework which pre-trains encoder and decoder simultaneously together with multi-level contrastive learning, our MACL outperforms all the existing SOTA baselines in almost all the metrics under different ratios of labeled data. Specifically, (1) when 10\% annotations are used, our proposed MACL gains a great improvement on all 4 multi-organ datasets and 4 ROI-based datasets: {\itshape e.g.} 1.72\% DSC higher than SOTA baseline PCL on ACDC, 7.87\% DSC and 8.61\% JC higher than SOTA baseline GCL on MMWHS, 1.48\% DSC higher than SOTA baseline PCL on CHAOS, 7.53\% DSC and 10.0\% JC higher than SOTA baseline GCL on Spleen.
Our proposed MACL also achieves the lowest (best) HD95, such as 5.23 on ACDC, 5.91 on MMWHS and 22.38 on HVSMR, 6.93 on Spleen and 21.53 on ISIC.
(2) When 25\% annotations are used, the performance improvement of our MACL against other SOTA baselines is narrowed. Because with more labeled samples, more supervisory information will be available, which results in that the improvement from self-supervised pre-training tends to saturate. Despite that, the gain of performance is still stable. {\itshape e.g.} 1.89\% DSC and 2.88\% JC higher than GCL on MMWHS, 1.45\% DSC and 3.42\% JC higher than GCL on Spleen.
And our proposed MACL achieves the best results of all metrics on all 4 multi-organ datasets, {\itshape e.g.} 83.68\% DSC, 72.47\% JC, 17.03 HD95 and 4.14 ASD on HVSMR, 79.88\% DSC, 67.45\% JC, 8.07 HD95 and 3.05 ASD. 
(3) Moreover, our proposed MACL pre-training framework can realize our initial goal effectively: pre-train a model with unlabeled data from upstream datasets and limited labeled data from downstream datasets to achieve comparative performance with the fully-supervised trained model. Specifically, the performance of our proposed MACL with 25\% annotations can be paralleled with that of Scratch with 100\% annotations (fully-supervised training). {\itshape e.g.} 89.65\% vs. 92.07\% DSC and 1.20 vs. 1.07 ASD on ACDC; 83.68\% vs. 84.95\% DSC, 72.47\% vs. 74.33\% JC 4.14 vs. 3.56 ASD on HVSMR; 88.96\% vs. 90.11\% DSC, 80.12\% vs. 82.01\% JC, 17.61 vs. 15.71 HD95, 6.79 vs. 6.17 ASD on ISIC.

\noindent{\textbf{Visualization of multi-organ segmentation results:}}
As shown in Figure \ref{Vis_1} and \ref{Vis_2}, we visualize the exemplar qualitative results of our MACL and 11 other methods on ACDC, MMWHS, CHAOS and HVSMR. Compared with different methods, it can be seen that our proposed MACL achieves a superior performance with a higher MIoU and more precise predictions of substructures across other 11 methods. {\itshape e.g.} all the 11 other methods except for our proposed MACL, predict some wrong segmentation region (right ventricle/yellow region) with a much lower MIoU on ACDC. Or other methods have some less precise predictions compared with our proposed MACL, like right atrium/sky-blue region on MMWHS.

\begin{figure*}[t]
  \centering
  \includegraphics[width=0.8\textwidth]{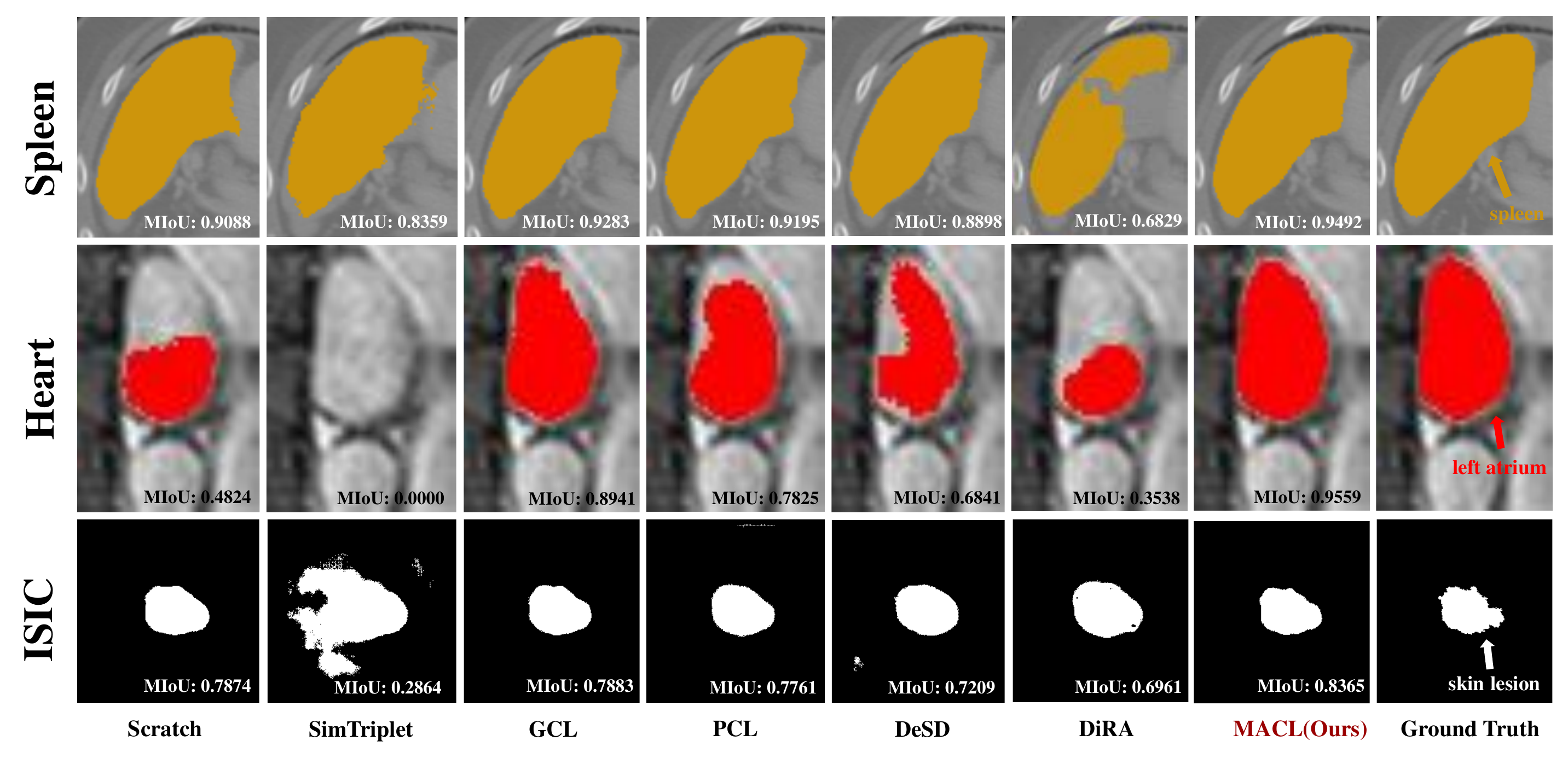}
  \caption{Visualization of ROI-based segmentation results on Spleen, Heart and ISIC datasets.}
  \label{Vis_ROI}
\end{figure*}

\noindent{\textbf{Visualization of ROI-based segmentation results:}}
Similar to multi-organ segmentation, we also present the visualization of ROI-based segmentation results on Spleen, Heart and ISIC in Figure \ref{Vis_ROI}.
We find clearly that our proposed MACL has a much more accurate and complete prediction segmentation region with a higher MIoU. Specifically, for left atrium/red region of Heart dataset, both DeSD and DiRA has a large missing segmentation region and SimTriplet even predicts nothing, while our proposed MACL has an impressive segmentation result with MIoU 0.9559.
\begin{table*}[t]
\centering
\caption{The ablation studies for each component of our proposed method.}
$\begin{aligned}
\Xhline{1.0pt}
\setlength{\tabcolsep}{3pt}
\begin{tabular}{lcccc|cccc|cccc}
\multicolumn{5}{c|}{\text { Settings }} & \multicolumn{4}{c|}{\text { ACDC (25\% label) }} & \multicolumn{4}{c}{\text {MMWHS (25\% label)}} \\
Framework & $\mathcal{L}_{g}$ & \itshape{decoder} & $\mathcal{L}_{ER}$ & $\mathcal{L}_{d}$ & DSC (\%) $\uparrow$ & JC (\%) $\uparrow$ & HD95 $\downarrow$ & ASD $\downarrow$ & DSC (\%) $\uparrow$ & JC (\%) $\uparrow$ & HD95 $\downarrow$ & ASD $\downarrow$ \\
\Xhline{1.0pt}
PCL & \checkmark & & & & 88.97 & 80.33 & 4.17 & 1.43 & 86.65 & 76.90 & 4.23 & 1.28 \\
PCL + decoder & \checkmark & \checkmark & & & 89.19 & 80.73 & 3.92 & 1.33 & 88.19 & 79.15 & 4.10 & 1.19 \\
MACL (image-level) & \checkmark & \checkmark & \checkmark & & 89.29 & 80.82 & 4.08 & 1.30 & 88.96 & 80.51 & 3.74 & ${\mathbf{0.94}}$ \\
\cellcolor{cyan!20!green!20!}MACL (image/pixel-level) & \cellcolor{cyan!20!green!20!}\checkmark & \cellcolor{cyan!20!green!20!}\checkmark & \cellcolor{cyan!20!green!20!}\checkmark & \cellcolor{cyan!20!green!20!}\checkmark & \cellcolor{cyan!20!green!20!}${\mathbf{89.64}}$ & \cellcolor{cyan!20!green!20!}${\mathbf{81.40}}$ & \cellcolor{cyan!20!green!20!}${\mathbf{3.59}}$ & \cellcolor{cyan!20!green!20!}${\mathbf{1.20}}$ & \cellcolor{cyan!20!green!20!}${\mathbf{89.90}}$ & \cellcolor{cyan!20!green!20!}${\mathbf{81.92}}$ & \cellcolor{cyan!20!green!20!}${\mathbf{3.68}}$ & 
\cellcolor{cyan!20!green!20!}$1.01$ \\
\Xhline{1.0pt}
\end{tabular}
\end{aligned}$
\label{Ablation}
\end{table*}

\begin{table*}[t]
\centering
\caption{The ablation studies of number of blocks of the decoder. $\lambda$ denotes scale factor used in downsampling.}
\begin{tabular}{c|cccc|cccc}
\Xhline{1.0pt}
& \multicolumn{4}{c|}{ACDC (10\% label)} & \multicolumn{4}{c}{MMWHS (10\% label)}  \\
Number of Blocks & DSC (\%) $\uparrow$ & JC (\%) $\uparrow$ & HD95 $\downarrow$ & ASD $\downarrow$ & DSC (\%) $\uparrow$ & JC (\%) $\uparrow$ & HD95 $\downarrow$ & ASD $\downarrow$ \\
\hline
0 & 84.91 & 74.16 & 6.54 & 2.65 & 65.98 & 52.32 & 8.07 & 3.16 \\
1 ($\lambda=0.5$) & 85.79 & 75.44 & 7.89 & 2.59 & 71.51 & 58.45 & 7.84 & 2.58 \\
2 ($\lambda=0.25$) & ${\mathbf{86.63}}$ & ${\mathbf{76.72}}$ & ${\mathbf{5.23}}$ & ${\mathbf{1.73}}$ & ${\mathbf{79.28}}$ & ${\mathbf{67.19}}$ & ${\mathbf{5.91}}$ & 1.98 \\
3 ($\lambda=0.125$) & 83.61 & 72.17 & 9.66 & 3.22 & 73.33 & 60.42 & 7.29 & 2.22 \\
4 ($\lambda=0.0625$) & 85.70 & 75.34 & 8.24 & 2.86 &75.57&62.79&6.39&${\mathbf{1.96}}$ \\
\Xhline{1.0pt} 
\end{tabular}
\label{block_numbers}
\end{table*}

\begin{table*}[t]
\centering
\caption{Comparison results of different connection modes of encoders and different stages of MACL.}
\begin{tabular}{c|cc|cc|cc}
\Xhline{1.0pt}
& \multicolumn{2}{c|}{\text { Settings }} & \multicolumn{2}{c|}{\text {ACDC}} & \multicolumn{2}{c}{\text {MMWHS}} \\
{Labeled\%} & Stages & Framework & DSC (\%) $\uparrow$ & HD95 $\downarrow$ & DSC (\%) $\uparrow$ & HD95 $\downarrow$ \\
\hline 
\multirow{5}*{10 \%} & \multirow{3}*{One Stage} & two independent encoders & 82.20 & 6.24 & 63.17 & 9.82 \\
& & encoders with EMA & 79.75 & 10.96 & 59.61 & 10.84 \\
& & two parameter-shared encoders &${\mathbf{86.63}}$ &${\mathbf{5.23}}$&${\mathbf{79.28}}$ &${\mathbf{5.91}}$ \\
\cline{2-7}
& \multirow{2}*{Two Stages} & stage one \textbf{w} downsample &81.15 &6.99 &60.13 &10.53 \\
& & stage one \textbf{w/o} downsample &81.46 &7.13 &59.73 &11.02 \\
\Xhline{1.0pt}

\multirow{5}*{25 \%} & \multirow{3}*{One Stage} & two independent encoders &89.09 &4.73 &86.68&4.14 \\
& & encoders with EMA &88.24 &4.94 &86.50 &4.49 \\
& & two parameter-shared encoders &${\mathbf{89.64}}$ &${\mathbf{3.59}}$&${\mathbf{89.90}}$ &${\mathbf{3.68}}$ \\
\cline{2-7}
& \multirow{2}*{Two Stages} & stage one \textbf{w} downsample &88.19&4.65&85.38&4.24 \\
& & stage one \textbf{w/o} downsample &89.28&3.80&84.60&5.01 \\
\Xhline{1.0pt}

\end{tabular}
\label{Connection Modes}
\end{table*}


\subsection{Ablation study}

\noindent{\textbf{Different parts of our proposed MACL:}} Ablation studies have been conducted to verify the effectiveness of each part of our proposed components, {\itshape e.g.} the decoder structure (introduce asymmetry), equivariant regularization $\mathcal{L}_{ER}$ and dense contrastive loss $\mathcal{L}_{d}$ of our proposed MACL framework.
Results are reported in Table \ref{Ablation}. It can be seen that each part of our proposed components contributes to the improvement of performance. 
Specifically, the introduction of decoder structure and pixel-level dense contrastive loss $\mathcal{L}_{d}$ improves the performance of our MACL framework to a greater extent, {\itshape i.e.} 1.54\% DSC and 2.25\% JC promotion on MMWHS when applying the decoder and 0.35\% DSC promotion on ACDC and 0.94\% DSC, 1.41\% JC promotion on MMWHS when applying $\mathcal{L}_{d}$. And these improvements verify the effectiveness of our proposed asymmetric CL structure and multi-level contrastive learning strategy.

\noindent{\textbf{Different numbers of blocks of decoder:}} We have also conducted an ablation study on the number of blocks (N = 0,1,2,3,4) of the decoder to verify the best number for choice.
As shown in Table \ref{block_numbers}, all the frameworks with a decoder outperforms that without a decoder, and the two-block decoder gets the best performance in almost all metrics except for ASD on MMWHS, which is used as the baseline setting for other experiments. Since we need to add downsampling with a scale factor according to the numbers of blocks of the decoder, the input size is perhaps too small when three/four blocks of the decoder, resulting in a damage to image details and thus degrading performance.

\begin{figure*}[t]
  \centering
  \includegraphics[width=0.9\textwidth]{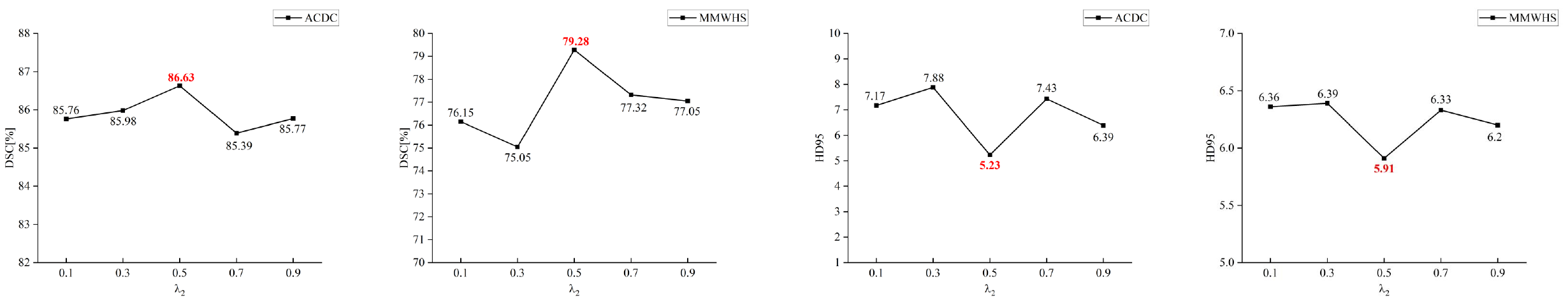}
  \caption{Comparison results of different $\lambda_2$ on ACDC and MMWHS with 10\% annotations. And $\lambda_2=0.5$ achieves the best which is used as baseline setting.}
  \label{lambda_2}
\end{figure*}

\noindent{\textbf{Ablation studies of $\lambda_2$:}}
According to Table \ref{Ablation}, we conclude that pixel-level dense CL loss $\mathcal{L}_{d}$ matters more than equivariant regularization $\mathcal{L}_{ER}$, therefore we conduct a detailed ablation experiment of the weighted term $\lambda_2$ of $\mathcal{L}_{d}$. From Figure \ref{lambda_2}, we conclude that our proposed pixel-level dense CL loss $\mathcal{L}_{d}$ has a local optimum within the range of 0.1 to 0.9, where $\lambda_2$ = 0.5 achieves the best performance, which is also used as our baseline setting. Specifically, when $\lambda_2$ = 0.5, our proposed MACL achieves the best DSC (86.63\% and 79.28\%) and HD95 (5.23 and 5.91) on both ACDC and MMWHS with 10\% annotations.

\begin{figure*}[t]
  \centering
  \includegraphics[width=\textwidth]{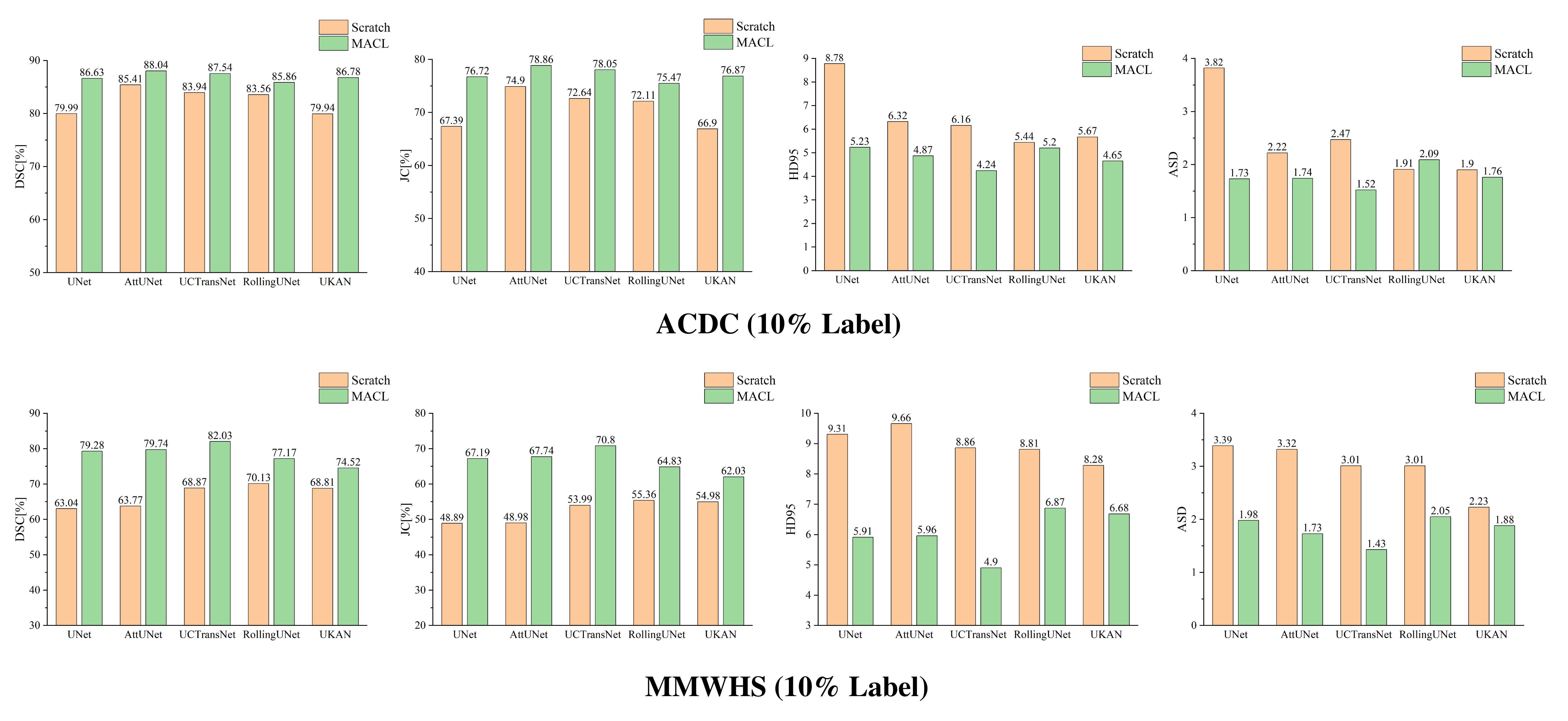}
  \caption{Comparison results of different variants of UNet backbone on ACDC and MMWHS. Recommend to enlarge the view for better visibility. The performance is better with higher DSC/JC and lower HD95/ASD. And the results demostrate our MACL can be generalized to different model backbones.}
  \label{backbone}
\end{figure*}

\noindent{\textbf{Different connection modes of encoders:}} Considering there exists different connection modes of encoders in these SOTA methods, such as parameter-shared encoders in SimCLR and SimSiam, and encoder with EMA in MoCo and BYOL. We explore the performance of different connection modes of the 
encoder between two branches. From Table \ref{Connection Modes}, we can conclude that the parameter-shared mode matters a lot especially in CL with negative sample pairs. Moreover, we further explore the performance between one-stage MACL (pre-train encoder and decoder simultaneously) and two-stage MACL (pre-train encoder and decoder 
separately). From Table \ref{Connection Modes}, we can see that our MACL in one-stage setting performs much better than MACL in two stage setting (The connection mode of encoders in Stage two of Two-stage MACL is parameter-shared).


\noindent{\textbf{Variant U-Net backbones:}} The decoders used for medical image segmentation are usually U-shape, and our MACL can also be generalized to other variants of U-Net. We have conducted experiments on different variants of U-Net to verify generalization and universality of our proposed MACL. Specifically, we choose AttUNet \cite{oktay2018attention} (decoder with attention blocks), UCTransNet \cite{wang2022uctransnet} (Transformer complement to CNN-based U-Net), RollingUNet \cite{Liu_Zhu_Liu_Yu_Chen_Gao_2024} (incorporate MLP into U-Net) and UKAN \cite{li2024ukanmakesstrongbackbone} (integrate Kolmogorov-Arnold Networks into U-Net) for comparison with the basic U-Net \cite{ronneberger2015u}. The results are shown in Figure \ref{backbone} and we can see that our MACL can be generalized to different model backbones, and our proposed MACL obtains an improvement among all 4 metrics on both ACDC and MMWHS.

\begin{figure*}[t]
  \centering
  \includegraphics[width=0.8\textwidth]{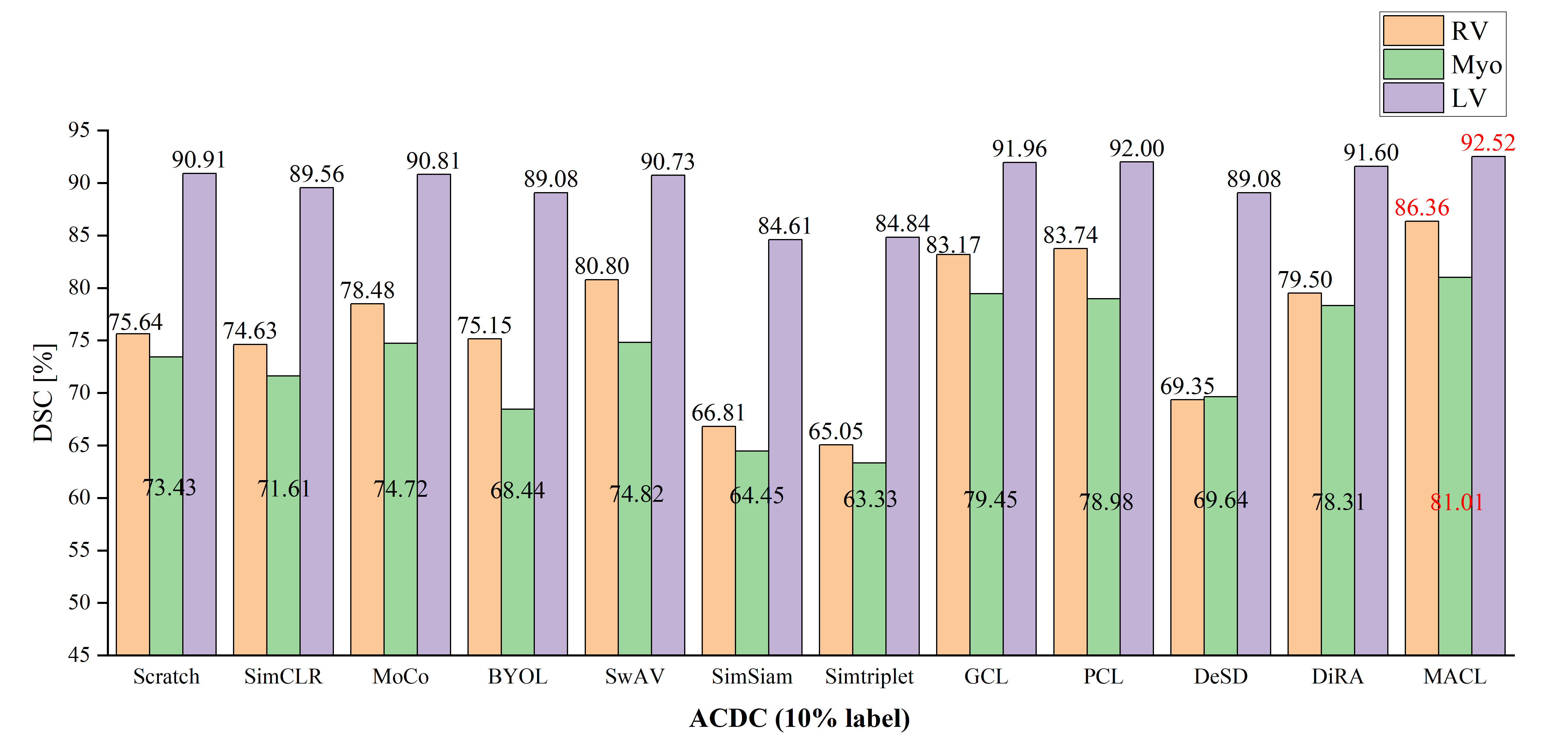}
  \caption{The performance (DSC) of each individual organ on ACDC with 10\% annotations.}
  \label{each_organ_ACDC}
\end{figure*}

\begin{figure*}[t]
  \centering
  \includegraphics[width=0.8\textwidth]{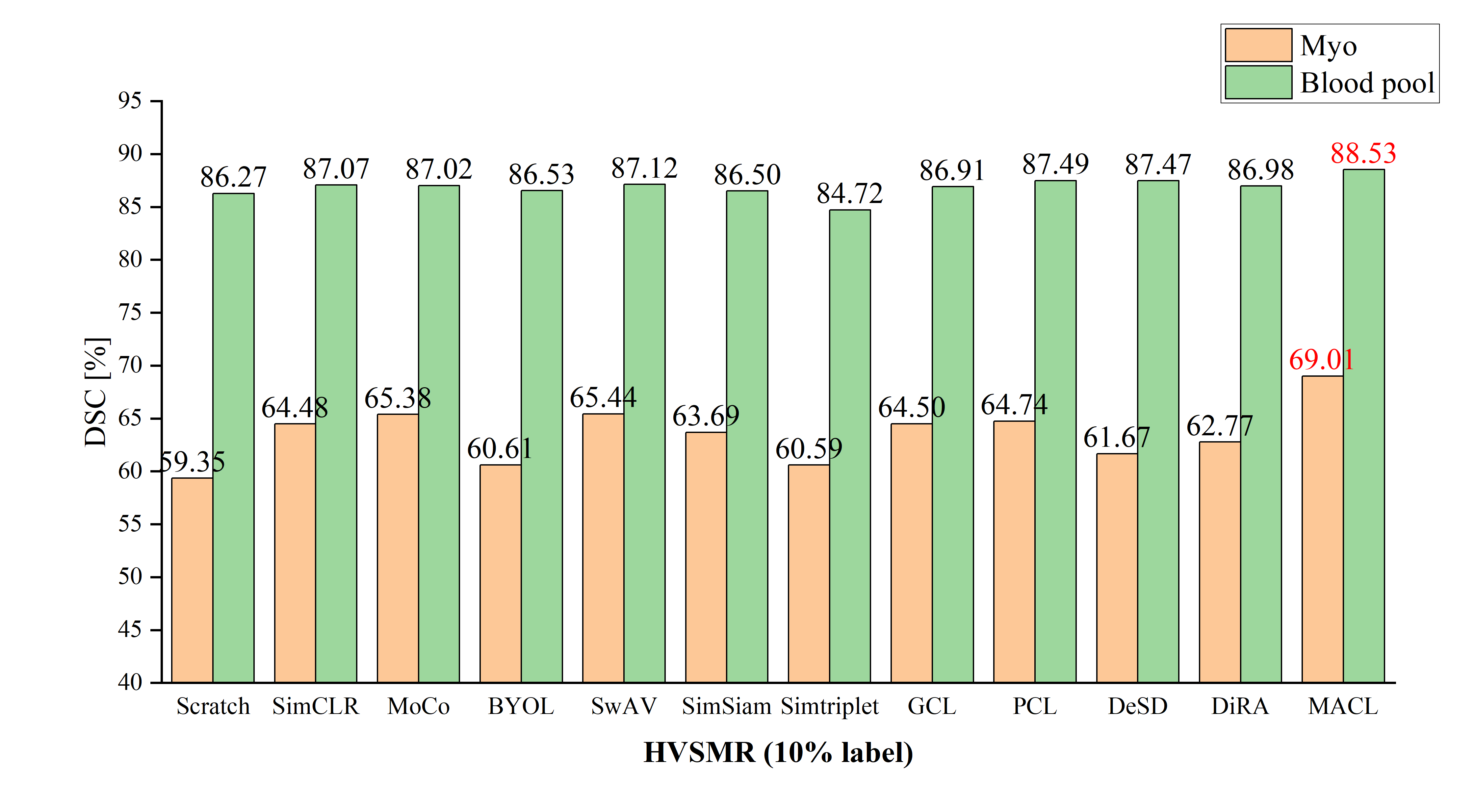}
  \caption{The performance (DSC) of each individual organ on HVSMR with 10\% annotations.}
  \label{each_organ_HVSMR}
\end{figure*}

\begin{figure*}[t]
  \centering
  \includegraphics[width=0.8\textwidth]{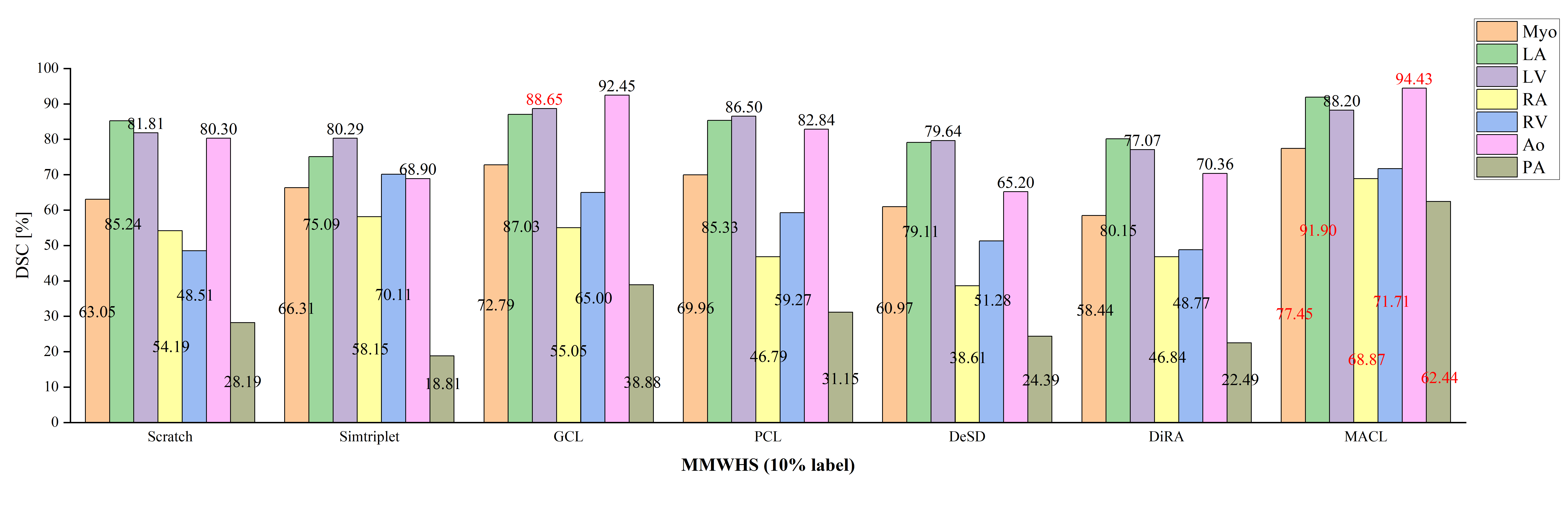}
  \caption{The performance (DSC) of each individual organ on MMWHS with 10\% annotations.}
  \label{each_organ_MMWHS}
\end{figure*}

\noindent{\textbf{Segmentation performance of each individual organ:}}
We also make a more detailed comparison of segmentation performance of each individual organ across different multi-organ datasets. As shown in Figure \ref{each_organ_ACDC}, \ref{each_organ_HVSMR} and \ref{each_organ_MMWHS},
our proposed MACL almost achieves all SOTA results (DSC) for individual organ segmentation on ACDC, HVSMR and MMWHS with 10\% annotations. Specifically, our proposed MACL gets SOTA DSC with RV (Right Ventricle) 86.36\%, Myo (Myocardium) 81.01\% and LV (Left Ventricle) 92.52\% on ACDC. And as for HVSMR, our MACL also achieves best results with Myo 69.01\% DSC and Blood pool 88.53\% DSC. Finally, for the multi-organ dataset with a larger number of categories (7 types) MMWHS, we also obtain a relatively satisfactory and convincing result: except for LV (a slightly lower than GCL, 88.20\% vs. 88.65\% DSC), our MACL achieves the highest DSC among all the other organs with a huge improvement compared with other baseline methods. For example, for PA (Pulmonary Artery), we get 62.44\% DSC, which is 23.56\% higher than SOTA baseline GCL. And for other 5 organ categories, compared with each SOTA baseline method, we can also get a promising improvement. 
\itshape{e.g.} LA: MACL 91.90\% vs. GCL 87.03\% (4.87\% $\uparrow$), RA: MACL 68.87\% vs. SimTriplet 58.15\% (10.72\% $\uparrow$).   
\section{Conclusion}\label{sec:abalation}
In this work, we propose a novel multi-level asymmetric CL framework named MACL for medical image segmentation pre-training. Specifically, we introduce an asymmetric CL structure to pre-train both encoder and decoder simultaneously in one stage to provide better initialization for segmentation models. Additionally, We develop a multi-level CL strategy that integrates correspondences across feature-level, image-level, and pixel-level projections to ensure encoders and decoders capture comprehensive details from representations of varying scales and granularities during the pre-training phase.
Experiments on multiple datasets indicate our MACL outperforms existing SOTA CL methods.

\bibliographystyle{IEEEtran}
\bibliography{TNNLS.bib}

\begin{thebibliography}{10}
\providecommand{\url}[1]{#1}
\csname url@samestyle\endcsname
\providecommand{\newblock}{\relax}
\providecommand{\bibinfo}[2]{#2}
\providecommand{\BIBentrySTDinterwordspacing}{\spaceskip=0pt\relax}
\providecommand{\BIBentryALTinterwordstretchfactor}{4}
\providecommand{\BIBentryALTinterwordspacing}{\spaceskip=\fontdimen2\font plus
\BIBentryALTinterwordstretchfactor\fontdimen3\font minus
  \fontdimen4\font\relax}
\providecommand{\BIBforeignlanguage}[2]{{%
\expandafter\ifx\csname l@#1\endcsname\relax
\typeout{** WARNING: IEEEtran.bst: No hyphenation pattern has been}%
\typeout{** loaded for the language `#1'. Using the pattern for}%
\typeout{** the default language instead.}%
\else
\language=\csname l@#1\endcsname
\fi
#2}}
\providecommand{\BIBdecl}{\relax}
\BIBdecl

\bibitem{chen2020simple}
T.~Chen, S.~Kornblith, M.~Norouzi, and G.~Hinton, ``A simple framework for
  contrastive learning of visual representations,'' in \emph{International
  conference on machine learning}.\hskip 1em plus 0.5em minus 0.4em\relax PMLR,
  2020, pp. 1597--1607.

\bibitem{he2020momentum}
K.~He, H.~Fan, Y.~Wu, S.~Xie, and R.~Girshick, ``Momentum contrast for
  unsupervised visual representation learning,'' in \emph{Proceedings of the
  IEEE/CVF conference on computer vision and pattern recognition}, 2020, pp.
  9729--9738.

\bibitem{zeng2021positional}
D.~Zeng, Y.~Wu, X.~Hu, X.~Xu, H.~Yuan, M.~Huang, J.~Zhuang, J.~Hu, and Y.~Shi,
  ``Positional contrastive learning for volumetric medical image
  segmentation,'' in \emph{International Conference on Medical Image Computing
  and Computer-Assisted Intervention}.\hskip 1em plus 0.5em minus 0.4em\relax
  Springer, 2021, pp. 221--230.

\bibitem{DeSD}
Y.~Ye, J.~Zhang, Z.~Chen, and Y.~Xia, ``Desd: Self-supervised learning
  with deep self-distillation for 3d medical image segmentation,'' in
  \emph{Medical Image Computing and Computer Assisted Intervention -- MICCAI
  2022}, L.~Wang, Q.~Dou, P.~T. Fletcher, S.~Speidel, and S.~Li, Eds.\hskip 1em
  plus 0.5em minus 0.4em\relax Cham: Springer Nature Switzerland, 2022, pp.
  545--555.

\bibitem{DiRA}
F.~Haghighi, M.~R.~H. Taher, M.~B. Gotway, and J.~Liang, ``Dira:
  Discriminative, restorative, and adversarial learning for self-supervised
  medical image analysis,'' in \emph{Proceedings of the IEEE/CVF Conference on
  Computer Vision and Pattern Recognition (CVPR)}, June 2022, pp.
  20\,824--20\,834.

\bibitem{liu2021simtriplet}
Q.~Liu, P.~C. Louis, Y.~Lu, A.~Jha, M.~Zhao, R.~Deng, T.~Yao, J.~T. Roland,
  H.~Yang, S.~Zhao \emph{et~al.}, ``Simtriplet: Simple triplet representation
  learning with a single gpu,'' in \emph{Medical Image Computing and Computer
  Assisted Intervention--MICCAI 2021: 24th International Conference,
  Strasbourg, France, September 27--October 1, 2021, Proceedings, Part II
  24}.\hskip 1em plus 0.5em minus 0.4em\relax Springer, 2021, pp. 102--112.

\bibitem{GCL}
K.~Chaitanya, E.~Erdil, N.~Karani, and E.~Konukoglu, ``Contrastive learning of
  global and local features for medical image segmentation with limited
  annotations,'' \emph{Advances in Neural Information Processing Systems},
  vol.~33, pp. 12\,546--12\,558, 2020.

\bibitem{long2015fully}
J.~Long, E.~Shelhamer, and T.~Darrell, ``Fully convolutional networks for
  semantic segmentation,'' in \emph{Proceedings of the IEEE conference on
  computer vision and pattern recognition}, 2015, pp. 3431--3440.

\bibitem{ronneberger2015u}
O.~Ronneberger, P.~Fischer, and T.~Brox, ``U-net: Convolutional networks for
  biomedical image segmentation,'' in \emph{Medical image computing and
  computer-assisted intervention--MICCAI 2015: 18th international conference,
  Munich, Germany, October 5-9, 2015, proceedings, part III 18}.\hskip 1em plus
  0.5em minus 0.4em\relax Springer, 2015, pp. 234--241.

\bibitem{chen2018encoder}
L.-C. Chen, Y.~Zhu, G.~Papandreou, F.~Schroff, and H.~Adam, ``Encoder-decoder
  with atrous separable convolution for semantic image segmentation,'' in
  \emph{Proceedings of the European conference on computer vision (ECCV)},
  2018, pp. 801--818.

\bibitem{oktay2018attention}
O.~Oktay, J.~Schlemper, L.~L. Folgoc, M.~Lee, M.~Heinrich, K.~Misawa, K.~Mori,
  S.~McDonagh, N.~Y. Hammerla, B.~Kainz \emph{et~al.}, ``Attention u-net:
  Learning where to look for the pancreas,'' \emph{arXiv preprint
  arXiv:1804.03999}, 2018.

\bibitem{wang2022uctransnet}
H.~Wang, P.~Cao, J.~Wang, and O.~R. Zaiane, ``Uctransnet: rethinking the skip
  connections in u-net from a channel-wise perspective with transformer,'' in
  \emph{Proceedings of the AAAI conference on artificial intelligence},
  vol.~36, no.~3, 2022, pp. 2441--2449.

\bibitem{Liu_Zhu_Liu_Yu_Chen_Gao_2024}
\BIBentryALTinterwordspacing
Y.~Liu, H.~Zhu, M.~Liu, H.~Yu, Z.~Chen, and J.~Gao, ``Rolling-unet:
  Revitalizing mlp’s ability to efficiently extract long-distance
  dependencies for medical image segmentation,'' \emph{Proceedings of the AAAI
  Conference on Artificial Intelligence}, vol.~38, no.~4, pp. 3819--3827, Mar.
  2024. [Online]. Available:
  \url{https://ojs.aaai.org/index.php/AAAI/article/view/28173}
\BIBentrySTDinterwordspacing

\bibitem{liu2024kan}
Z.~Liu, Y.~Wang, S.~Vaidya, F.~Ruehle, J.~Halverson, M.~Solja{\v{c}}i{\'c},
  T.~Y. Hou, and M.~Tegmark, ``Kan: Kolmogorov-arnold networks,'' \emph{arXiv
  preprint arXiv:2404.19756}, 2024.

\bibitem{li2024ukanmakesstrongbackbone}
\BIBentryALTinterwordspacing
C.~Li, X.~Liu, W.~Li, C.~Wang, H.~Liu, Y.~Liu, Z.~Chen, and Y.~Yuan, ``U-kan
  makes strong backbone for medical image segmentation and generation,'' 2024.
  [Online]. Available: \url{https://arxiv.org/abs/2406.02918}
\BIBentrySTDinterwordspacing

\bibitem{isensee2021nnu}
F.~Isensee, P.~F. Jaeger, S.~A. Kohl, J.~Petersen, and K.~H. Maier-Hein,
  ``nnu-net: a self-configuring method for deep learning-based biomedical image
  segmentation,'' \emph{Nature methods}, vol.~18, no.~2, pp. 203--211, 2021.

\bibitem{oord2018representation}
A.~v.~d. Oord, Y.~Li, and O.~Vinyals, ``Representation learning with
  contrastive predictive coding,'' \emph{arXiv preprint arXiv:1807.03748},
  2018.

\bibitem{caron2020unsupervised}
M.~Caron, I.~Misra, J.~Mairal, P.~Goyal, P.~Bojanowski, and A.~Joulin,
  ``Unsupervised learning of visual features by contrasting cluster
  assignments,'' \emph{Advances in neural information processing systems},
  vol.~33, pp. 9912--9924, 2020.

\bibitem{grill2020bootstrap}
J.-B. Grill, F.~Strub, F.~Altch{\'e}, C.~Tallec, P.~Richemond, E.~Buchatskaya,
  C.~Doersch, B.~Avila~Pires, Z.~Guo, M.~Gheshlaghi~Azar \emph{et~al.},
  ``Bootstrap your own latent-a new approach to self-supervised learning,''
  \emph{Advances in neural information processing systems}, vol.~33, pp.
  21\,271--21\,284, 2020.

\bibitem{chen2021exploring}
X.~Chen and K.~He, ``Exploring simple siamese representation learning,'' in
  \emph{Proceedings of the IEEE/CVF conference on computer vision and pattern
  recognition}, 2021, pp. 15\,750--15\,758.

\bibitem{hervella2020multi}
{\'A}.~S. Hervella, L.~Ramos, J.~Rouco, J.~Novo, and M.~Ortega, ``Multi-modal
  self-supervised pre-training for joint optic disc and cup segmentation in eye
  fundus images,'' in \emph{ICASSP 2020-2020 IEEE international conference on
  acoustics, speech and signal processing (ICASSP)}.\hskip 1em plus 0.5em minus
  0.4em\relax IEEE, 2020, pp. 961--965.

\bibitem{wang2020self}
Y.~Wang, J.~Zhang, M.~Kan, S.~Shan, and X.~Chen, ``Self-supervised equivariant
  attention mechanism for weakly supervised semantic segmentation,'' in
  \emph{Proceedings of the IEEE/CVF Conference on Computer Vision and Pattern
  Recognition}, 2020, pp. 12\,275--12\,284.

\bibitem{PCL23}
X.~Xu, T.~Wang, Y.~Shi, H.~Yuan, Q.~Jia, M.~Huang, and J.~Zhuang, ``Whole heart
  and great vessel segmentation in congenital heart disease using deep neural
  networks and graph matching,'' in \emph{Medical Image Computing and Computer
  Assisted Intervention--MICCAI 2019: 22nd International Conference, Shenzhen,
  China, October 13--17, 2019, Proceedings, Part II 22}.\hskip 1em plus 0.5em
  minus 0.4em\relax Springer, 2019, pp. 477--485.

\bibitem{ACDC}
O.~Bernard, A.~Lalande, C.~Zotti, F.~Cervenansky, X.~Yang, P.-A. Heng,
  I.~Cetin, K.~Lekadir, O.~Camara, M.~A. Gonzalez~Ballester, G.~Sanroma,
  S.~Napel, S.~Petersen, G.~Tziritas, E.~Grinias, M.~Khened, V.~A. Kollerathu,
  G.~Krishnamurthi, M.-M. Rohé, X.~Pennec, M.~Sermesant, F.~Isensee,
  P.~Jäger, K.~H. Maier-Hein, P.~M. Full, I.~Wolf, S.~Engelhardt, C.~F.
  Baumgartner, L.~M. Koch, J.~M. Wolterink, I.~Išgum, Y.~Jang, Y.~Hong,
  J.~Patravali, S.~Jain, O.~Humbert, and P.-M. Jodoin, ``Deep learning
  techniques for automatic mri cardiac multi-structures segmentation and
  diagnosis: Is the problem solved?'' \emph{IEEE Transactions on Medical
  Imaging}, vol.~37, no.~11, pp. 2514--2525, 2018.

\bibitem{PCL24}
X.~Zhuang, ``Challenges and methodologies of fully automatic whole heart
  segmentation: a review,'' \emph{Journal of healthcare engineering}, vol.~4,
  no.~3, pp. 371--407, 2013.

\bibitem{PCL16}
D.~F. Pace, A.~V. Dalca, T.~Geva, A.~J. Powell, M.~H. Moghari, and P.~Golland,
  ``Interactive whole-heart segmentation in congenital heart disease,'' in
  \emph{Medical Image Computing and Computer-Assisted Intervention--MICCAI
  2015: 18th International Conference, Munich, Germany, October 5-9, 2015,
  Proceedings, Part III 18}.\hskip 1em plus 0.5em minus 0.4em\relax Springer,
  2015, pp. 80--88.

\bibitem{CHAOS2021}
\BIBentryALTinterwordspacing
A.~E. Kavur, N.~S. Gezer, M.~Barış, S.~Aslan, P.-H. Conze, V.~Groza, D.~D.
  Pham, S.~Chatterjee, P.~Ernst, S.~Özkan, B.~Baydar, D.~Lachinov, S.~Han,
  J.~Pauli, F.~Isensee, M.~Perkonigg, R.~Sathish, R.~Rajan, D.~Sheet,
  G.~Dovletov, O.~Speck, A.~Nürnberger, K.~H. Maier-Hein, G.~{Bozdağı Akar},
  G.~Ünal, O.~Dicle, and M.~A. Selver, ``{CHAOS Challenge - combined (CT-MR)
  healthy abdominal organ segmentation},'' \emph{Medical Image Analysis},
  vol.~69, p. 101950, Apr. 2021. [Online]. Available:
  \url{http://www.sciencedirect.com/science/article/pii/S1361841520303145}
\BIBentrySTDinterwordspacing

\bibitem{bakas2018identifying}
S.~Bakas, M.~Reyes, A.~Jakab, S.~Bauer, M.~Rempfler, A.~Crimi, R.~T. Shinohara,
  C.~Berger, S.~M. Ha, M.~Rozycki \emph{et~al.}, ``Identifying the best machine
  learning algorithms for brain tumor segmentation, progression assessment, and
  overall survival prediction in the brats challenge,'' \emph{arXiv preprint
  arXiv:1811.02629}, 2018.

\bibitem{heller2021state}
N.~Heller, F.~Isensee, K.~H. Maier-Hein, X.~Hou, C.~Xie, F.~Li, Y.~Nan, G.~Mu,
  Z.~Lin, M.~Han \emph{et~al.}, ``The state of the art in kidney and kidney
  tumor segmentation in contrast-enhanced ct imaging: Results of the kits19
  challenge,'' \emph{Medical image analysis}, vol.~67, p. 101821, 2021.

\bibitem{antonelli2022medical}
M.~Antonelli, A.~Reinke, S.~Bakas, K.~Farahani, A.~Kopp-Schneider, B.~A.
  Landman, G.~Litjens, B.~Menze, O.~Ronneberger, R.~M. Summers \emph{et~al.},
  ``The medical segmentation decathlon,'' \emph{Nature communications},
  vol.~13, no.~1, p. 4128, 2022.

\bibitem{DBLP:journals/corr/abs-1902-03368}
\BIBentryALTinterwordspacing
N.~C.~F. Codella, V.~Rotemberg, P.~Tschandl, M.~E. Celebi, S.~W. Dusza, D.~A.
  Gutman, B.~Helba, A.~Kalloo, K.~Liopyris, M.~A. Marchetti, H.~Kittler, and
  A.~Halpern, ``Skin lesion analysis toward melanoma detection 2018: {A}
  challenge hosted by the international skin imaging collaboration {(ISIC)},''
  \emph{CoRR}, vol. abs/1902.03368, 2019. [Online]. Available:
  \url{http://arxiv.org/abs/1902.03368}
\BIBentrySTDinterwordspacing

\end{thebibliography}

\end{document}